\documentclass[a4paper,fleqn]{cas-dc}
\usepackage{graphicx}
\usepackage{epstopdf}  
\usepackage{amsfonts,amssymb}
\usepackage{amsmath}
\usepackage{subcaption}
\usepackage{algorithm}
\usepackage{algorithmic}
\usepackage{multirow, booktabs}
\usepackage{cite}
\usepackage{hyperref}
\usepackage{bookmark} 
\usepackage{indentfirst}

\usepackage[numbers]{natbib}
\usepackage{booktabs}

\begin{document}
\begin{sloppypar}
\def\floatpagepagefraction{1}
\def\textpagefraction{.001}
\shorttitle{}
\shortauthors{Junchao Lin, Yuan Wan and Jingwen Xu et~al.}

\title [mode = title]{Semantic Graph Neural Network with Multi-measure Learning for Semi-supervised Classification}                      

\author[1]{Junchao Lin}[orcid=0000-0002-6173-4601]                    
\ead{linjunchao@whut.edu.cn}
\address[1]{Mathematical Department, School of Science, Wuhan University of Technology, Wuhan, 430070, China}

\author[1]{Yuan Wan}[style=chinese]
\ead{wanyuan@whut.edu.cn}
\cormark[1]

\author[1]{Jingwen Xu}[style=chinese]
\ead{jingwenxu@whut.edu.cn}

\author[2]{Xingchen Qi}[style=chinese]
\ead{xingchen.qi@utexas.edu}
\address[2]{Electrical and Computer Engineering Department, The University of Texas at Austin, Austin, 78712, Texas, US}

\cortext[cor1]{Corresponding author: Yuan Wan}

\begin{abstract}
	Graph Neural Networks (GNNs) have attracted increasing attention in recent years and have achieved excellent performance in semi-supervised node classification tasks. The success of most GNNs relies on one fundamental assumption, i.e., the original graph structure data is available.
	However, recent studies have shown that GNNs are vulnerable to the complex underlying structure of the graph, making it necessary to learn comprehensive and robust graph structures for downstream tasks, rather than relying only on the raw graph structure.
	In light of this, we seek to learn optimal graph structures for downstream tasks and propose a novel framework for semi-supervised classification. 
	Specifically, based on the structural context information of graph and node representations, we encode the complex interactions in semantics and generate semantic graphs to preserve the global structure. Moreover, we develop a novel multi-measure attention layer to optimize the similarity rather than prescribing it a priori, so that the similarity can be adaptively evaluated by integrating measures. These graphs are fused and optimized together with GNN towards semi-supervised classification objective.
	Extensive experiments and ablation studies on six real-world datasets clearly demonstrate the effectiveness of our proposed model and the contribution of each component.

\end{abstract}

\begin{keywords}
	Graph Neural Networks \sep Node Classification \sep Semantic Graph Learning \sep Multi-measure Learning
\end{keywords}

\maketitle

\section{Introduction}\label{sec1}
Graph networks, such as citation networks, social networks, and biology networks, are ubiquitous in the real world \cite{wu2020comprehensive}. Graph Neural Networks (GNNs), as powerful tools for learning deep representations in dealing with graph data, aim to learn node embeddings by aggregating and transforming the topological structure. The learned node embeddings are then used in semi-supervised classification tasks. GNNs have attracted considerable attention in tackling graph analysis problems and are widely used in node classification \cite{Mixhop, demonet}, visual reasoning \cite{chen2018iterative}, and recommender systems \cite{song2019session, ying2018graph, fan2019graph}. The advent of GNNs realizes the end-to-end learning mode of graph data and provides a highly competitive learning scheme for the tasks in many application scenarios of graph data.

In recent years, GNNs have been developed rapidly and achieved excellent performance in semi-supervised node classification tasks. Most of them usually follow a message-passing scheme where node embeddings are obtained by aggregating the feature information of the topological neighbors in each convolutional layer and the learned embeddings are then used in classification tasks. This fusion strategy learns the node embeddings based on topology and feature information to preserve good structure of the graph.
As a model for analyzing graph structured data and embedding, GNN is highly sensitive to the quality of graph structure and its utilization. However, insufficient messaging and the suboptimal provided graphs inevitably pose a great challenge to the application of GNNs to real-world problems.
On the one hand, the message-passing processing is dependent on a comprehensive graph structure. Since the underlying structure of the graph is very complex, both the local and global structures need to be considered \cite{10.1145/1553374.1553494}. 
However, most methods neglect to consider the global graph structure, and the limited structural information may result in inferior performance. Therefore, how to simultaneously preserve the local and global structure is a tough problem \cite{wang_structural_2016}.
On the other hand, recent studies have shown that GNNs are susceptible to the quality of graph structure \cite{chen2021understanding}. In reality, many graphs are inevitably noisy or incomplete. This results in the raw graph being suboptimal for downstream tasks over topological neighbors. It is necessary to construct feature graphs to solve this problem, while the measure of most methods is pre-defined, which inevitably leads to the lack of robustness of the graph structure.
How to adaptively unify measure rules, for robust graph structure learning and full utilization of feature information, is another main problem.
Therefore, learning the optimal node embeddings in GNNs still faces the following major challenges:

\noindent\textbf{Global Graph Structure Preserving.}
Due to the characteristics of GNNs, the majority of existing GNNs \cite{Mixhop, bo2021beyond, AMGCN} have been designed to observe the low-level neighborhood of nodes when generating node representations.
Although they preserve local structural information well and effectively alleviate over-smoothing, they also bring a problem that the depth of message passing is insufficient, so that the global graph structure information is unintentionally ignored. 
In other words, they are more concerned with the local structure preservation of a graph for knowledge embedding, resulting the global structure of the graph is not well preserved.
However, message-passing between different nodes relies on a comprehensive graph structure, and a large number of high-order relationships with diverse semantics should be considered. 
Therefore, we need to construct an information encoding module based on semantic information, which will fully consider the global graph structure information, rather than just the low-level local one.

\noindent \textbf{Graph Structure Robustness Enhancing.}
Since the raw topology graph structure is susceptible to noise or unnoticeable perturbation, the performance degrades substantially. Without the support of label information, most GNNs address the problem by graph construction, i.e., optimizing graphs with some pre-defined rules and features.
A common strategy of graph construction is to first calculate the similarity between pairs of nodes based on one pre-defined measure, and then use the similarity to construct graphs for the downstream tasks. 
However, multiple criteria \cite{HGSL, AMGCN, zhang2019unsupervised} used to estimate the similarity can lead to different representations of the pairwise relations, and decisions about measure will lead to bias in the model and underutilization of feature information.
Therefore, we should consider to construct a multi-measure learning method for GNN that adaptively fuses multiple measure criteria for a robust graph structure, thereby narrowing the gap between the extracted graph and the optimal graph for downstream tasks.

Recently, attentional mechanisms have become one of the most influential mechanisms in deep learning \cite{Vaswani2017Attention} and have proven their effectiveness in GNNs \cite{GAT, SGAT, KGAT}. By observing a sequence of inputs, the attentional mechanism can decide which parts of the inputs to look at in order to gather the most useful information for output. 
Given their excellent information fusion capability, GNNs should be able to extract and fuse the most correlated information for classification tasks. However, the major obstacle is that information transfer between graph data ignores global structure preservation and leads to the insufficient feature information utilization. 
In order to embed graph knowledge sufficiently, can we design a new framework based on GNN that can not only consider the comprehensive structure of graphs, but also address the weakness of incomplete use of feature information and make the graph structure more robust?

To address these issues, a novel method named \textbf{S}emantic \textbf{G}raph \textbf{N}eural \textbf{N}etwork with \textbf{M}ulti-measure \textbf{L}earning (ML-SGNN) is proposed for semi-supervised classification. Based on the structural context information of graph and node representations, we encode the complex interactions in semantics and generate semantic graphs to preserve the global structure of the graph.
Furthermore, ML-SGNN utilizes the attention mechanism to capture sufficient feature information of the graph. 
Finally, based on the topology information, semantic information and features fusion information, the node embeddings are acquired and used for classification tasks and ${{l}_{2,1}}$-regularizer to reduce the generalization error of the model.

Technically, in order to fully preserve the global graph structure, we design a semantic information coding module based on random walk to extract higher-order semantic information in the graph structure. Specifically, we first randomly extract a large number of fixed-length sentences in the graph, where the sampling process of the tail node is optimized. Then we extract the semantic information by calculating the semantic similarity between node pairs, and define an embeddings layer of relevant semantic information based on the frequency matrix to preserve the global graph structure.
In addition, we construct a multi-measure learning method to extract the most correlated information for bridging the gap between the extracted graph and the best graph for downstream tasks. Specifically, we use the attention mechanism to automatically learn the importance weights of different measures, thereby adaptively fusing the multi-measure feature graph, and input the fused robust feature graph into the embedding layer. In this way, node labels can supervise the learning process and adaptively enhance graph structure robustness.
The major contributions of this work are briefly summarized below:
\begin{itemize}
	\item	Most existing GNNs are designed to observe the low-level neighborhood of nodes, resulting in lacking full consideration of the global structure of the graph. For this purpose, we innovatively construct a semantic information encoding module and propose a novel framework ML-SGNN.
	
	\item To address the problem of insufficient graph structure robustness caused by pre-defined measure rules, ML-SGNN adaptively fuses feature subgraphs constructed by multi-measures via attention mechanism.
	
	\item Extensive experiments and ablation studies have been conducted on six real datasets to validate the effectiveness of ML-SGNN against the state-of-the-art GNNs and the contribution of each component. The experiments show that it can adaptively adjust the weights to obtain an optimal structure and extract the most relevant feature embeddings of the graph.
\end{itemize}

The remaining part of this paper is organized as follows. We introduced the related works in Section \ref{sec2} and proposed our methods in Section \ref{sec3}. Experiments are shown in Section \ref{sec4} and we concluded this work in Section \ref{sec5}.

\begin{figure*}
	\centering
	\includegraphics[width=0.9\textwidth]{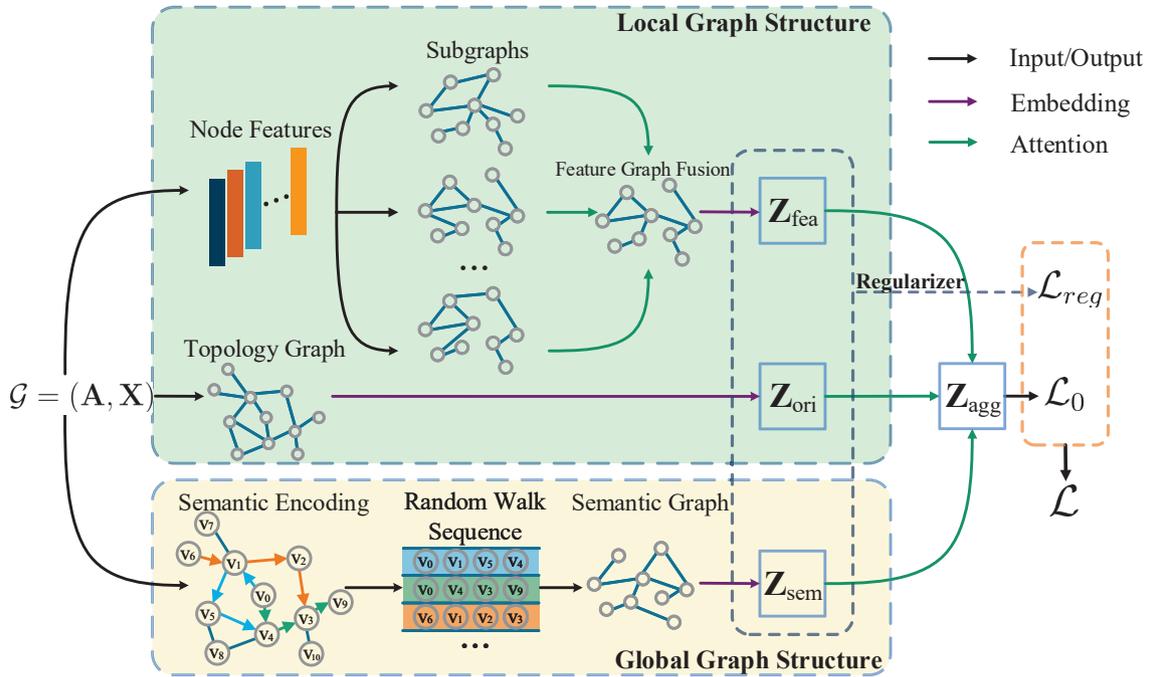}
	\caption{The framework of ML-SGNN model, consists of three GNN modules designed for node embedding, two attention layers used for graph fusion, and an optimizer used for improve generalization, where feature subgraphs are constructed by multi-measures and semantic graph is constructed by optimized semantic encoding to preserve the local and global structure. More details are described in Section \ref{sec3}.}
	\label{flow_chart}
\end{figure*}

\section{Related Work}\label{sec2}
\subsection{Graph Neural Network (GNN)}
\citet{GCN} simplify ChebNet \cite{ChebNet} by using a first-order approximation, leading to what we know as Graph Convolutional Neural Network. In recent years, GNN has attracted much interest as an emerging technology for learning graph data. Based on this, a generic degree-specific graph neural network called DEMO-Net has been proposed to recursively identify 1-hop neighborhood structures \cite{demonet}. \citet{MoNet} provide a generic spatial domain framework called MoNet for deep learning on non-Euclidean domains such as graphs and manifolds. Mixhop \cite{Mixhop} learns node representations by simultaneously aggregating the first-order and higher-order neighbor features of each layer. \citet{IDGL} believe that the original graphs after feature extraction and transformation may not reflect the "true" topology of the graph, and provide IDGL which uses adaptive graph regularizer to control the quality of the learning graph. On this basis, AM-GCN \cite{AMGCN} adaptively extracts the correlated information from node features and topological structures for classification.

\subsection{Attentional Graph Convolutional Network}
Attention mechanisms, which have become a research trend in deep learning, deal with variable-sized data and encourage the model to focus on the most salient parts of the data \cite{Vaswani2017Attention}. GAT \cite{GAT} uses the attention mechanism to define graph convolution. SGAT \cite{SGAT} is proposed to learn sparse attention coefficients under a $l_0$-norm regularizer, and the learned sparse attentions are then utilized for edge-sparsified graph. KGAT \cite{KGAT} explicitly models the high-order connectivities in knowledge graph in an end-to-end method. HGAN \cite{HGAN} generates node embedding by aggregating features from meta-path neighbors based on hierarchical attention. SPAGAN \cite{SPAGAN} is proposed to explore high-order path-based attentions within a layer. However, the above methods do not fully consider the information of node features and the case of multi-measures when embedding the nodes, which may not fully exploit the rich information in the data.

\subsection{Graph Embedding Method based on Random Walk}
To preserve the graph structure, node embeddings that occur in similar contexts tend to be given the same labels. For this purpose, GNNs map nodes or graphs to a low-dimensional space, which is called graph embedding, and some of them can be classified as random walk based graph embedding method \cite{goyal2018graph, hamilton2018representation, yang2015network, li2017attributed}.

Random walk plays an important role in semi-supervised learning for a variety of problems \cite{2016Revisiting}. Deepwalk \cite{deepwalk} learns the social representation of a network through truncated random walk, which can also achieve better results when there are few labeled nodes in the network. On this basis, GraRep \cite{GraRep} considers its higher-order transition probability derived from the random walk. 
DNGR proposed by \citet{DNGR} adopt a random surfing model to capture graph structural information directly.
To explore the global graph structure, \citet{li2018deeper} propose random walk models for training GCN. \citet{DGCN} introduce a dual graph convolution structure to capture information about the coexistence of nodes through random walks sampled from a graph. To learn more about graph embedding methods based on random walk, please refer to the related articles \cite{qiu2018network, jin2017predicting}. In our method, we encode the semantic similarity between nodes based on random walk to preserve the global structure.

\section{Approach}\label{sec3}
The overall architecture of the model is shown in Fig. \ref{flow_chart}. The feature subgraphs are first created by multi-measure. Then, an attention layer is used to obtain feature graph fusion, where two graph embedding modules are used to embed the fused feature graph and the original topology graph for local structure preservation. Subsequently, the semantic graph is constructed based on semantic information encoding, and another embedding module is used to embed the semantic graph for global structure preservation. Finally, for semi-supervised classification tasks, we aggregate the above three embeddings and obtain the most relevant embedding for supervised learning. In parallel, we introduce the three embeddings into the sparsity-inducing ${{l}_{2,1}}$ regularizer for unsupervised learning.

Here we present the notations used in this paper. Given a set of data points $\mathcal{X}=\{{{\mathbf{x}}_{1}},\ldots ,{{\mathbf{x}}_{l}},{{\mathbf{x}}_{l+1}},\ldots ,{{\mathbf{x}}_{n}}\}$ and a set of labels $C=\{1,\ldots ,c\}$, the first $l$ points have labels $\{{{y}_{1}},\ldots, {{y}_{l}}\}\in C$ and the remaining points are unlabeled. We focus on semi-supervised classification in graph $\mathcal{G}=({{\mathbf{A}}_{\text{ori}}},\mathbf{X})$, where ${{\mathbf{A}}_{\text{ori}}}\in {{\mathbb{R}}^{n\times n}}$ is the original graph adjacency matrix with $n$ nodes and $\mathbf{X}\in {{\mathbb{R}}^{n\times d}}$ is the node feature matrix with $d$ dimension. Specially, the ${{l}_{2,1}}$-norm of $\mathbf{Z}\in {{\mathbb{R}}^{n\times m}}$ is defined as $\|\mathbf{Z}{\|_{2,1}}=\sum\nolimits_{i=1}^{n}{\sqrt{\sum\nolimits_{j=1}^{m}{\mathbf{Z}_{i,j}^{2}}}}=\sum\nolimits_{i=1}^{n}{\|{{\mathbf{Z}}_{i,:}}{{\|}_{2}}}$.

\subsection{Multi-Measure Learning for Local Structure}\label{3.1}
In the scenario where no labeling information is available, a common strategy of most GNNs is to directly utilize some criteria based on prior knowledge to compute the similarity, and then the constructed similarity graph is consumed in a downstream task. Obviously, the evaluation of similarity between nodes is subjective. This poses challenges to the existing GNNs since most of their similarity measures are fixed and single. Some works \cite{YUAN2022194, IDGL} consider multi-measures, while overmuch adjustable parameters lead to poor interpretability. Motivated by addressing this issue, we design an adaptive multi-measure method for learning graph embeddings trained jointly with a task-dependent GNN model.

\subsubsection{\textbf{Multi-Measure Learning}}
For the full information of the node features in the feature space, multiple feature graphs $\mathcal{G}_{\text{fea}}^{q}=(\mathbf{A}^{q},\mathbf{X})$ are constructed based on the node feature matrix $\mathbf{X}$ and chosen measure functions ${{f}^{q}}$, $q=1\ldots Q$, where $\mathbf{A}^{q}$ is the adjacency matrix. In order to ensure that the information of the similarity matrix extracted by the feature is more abundant and multi perspective, three measure functions including
Cosine function \cite{HGSL} ${{f}_{1}}={\mathbf{x}_{i}^{\text{T}}{{\mathbf{x}}_{j}}}/{\| {{\mathbf{x}}_{i}}{{\| }_{2}}\| {{\mathbf{x}}_{j}}{{\| }_{2}}}\;$, Gaussian function \cite{AMGCN} ${{f}_{2}}=\exp (\| {{\mathbf{x}}_{i}}-{{\mathbf{x}}_{j}}\| _{2}^{2}2/t)$,
and a novel parameter free similarity measure with steerable sparsity (named Sparsity function), which has been derivated in \cite{zhang2019unsupervised} 
${f}_{3}=\left(\mathbf{e}_{i(k+1)}-\mathbf{e}_{i j} / k \mathbf{e}_{i(k+1)}-\sum_{j=1}^{k} \mathbf{e}_{i j}\right)_{+}$ where $\mathbf{e}_{i j}=\left\|\mathbf{x}_{i}-\mathbf{x}_{j}\right\|_{2}^{2}$,
are chosen to utilize.

We first calculate the similarity matrices ${{\mathbf{S}}^{q}}={{f}^{q}}(\mathbf{X})\in {{\mathbb{R}}^{n\times n}}$. In order to reduce the computational cost in the processing of preserving the local structure, we proceed to extract symmetric sparse adjacency matrix $\mathbf{A}^{q}$ from ${\mathbf{S}}^{q}$ by masking off those elements in ${\mathbf{S}}_{i}^{q}$ which are relatively further from node $i$.
\begin{equation}
\mathbf{A}_{i,j}^{q}=\left\{ \begin{matrix}
1 & \mathbf{S}_{i,j}^{q}\in \Phi_{i}^{q}  \\
0 & \text{otherwise,}  \\
\end{matrix} \right.
\end{equation}
where $\Phi_{i}^{q}$ denotes the $k$-Nearest Neighbor for each node $i$, and parameter $k$ is preset.

\subsubsection{\textbf{Multi-graph Attention Layer}}
By the above calculation, feature subgraphs can be obtained by node features. However, to reduce the reference of hyper-parameters and ensure the rationality of the model, a multi-graph attention layer needs to be introduced when merging multiple subgraphs.

The overall feature graph, denoted as ${{\mathbf{A}}_{\text{fea}}}$, can be obtained by fusing subgraph $\mathbf{A}^{q}$ through a multi-graph attention layer \cite{yun2019graph}:

\begin{equation} \label{psi_measures}
{{\mathbf{A}}_{\text{fea}}}={{\Psi }_{\text{\text{fea}}}}(\underset{q=1}{\overset{Q}\|}\,[\mathbf{A}^{q}]),
\end{equation}
where $\underset{q=1}{\overset{Q}\|}\,[\mathbf{A}_{}^{q}]\in {{\mathbb{R}}^{n\times n\times Q}}$ is the stacked matrix of the candidate feature subgraphs and $\mathop{||}$ represents concatenation, ${{\Psi }_{\text{fea}}}$ denotes an attention layer. Specifically, we first use a single layer perception layer to learn the above stacked matrix:
\begin{equation}
\text{SLP}(\underset{q=1}{\overset{Q}\|}\,[\mathbf{A}_{\text{fea}}^{q}])=\sigma ({{\mathbf{W}}_{\text{fea}}}\cdot {{\underset{q=1}{\overset{Q}\|}\,[\mathbf{A}_{\text{fea}}^{q}]^{T}}}+{{\mathbf{b}}_{\text{fea}}})
\end{equation}
where ${{\mathbf{W}}_{\text{fea}}}\in {{\mathbb{R}}^{h\times n}}$ and ${{\mathbf{b}}_{\text{fea}}}\in {{\mathbb{R}}^{h\times n}}$ are the shared weight matrix and bias vector of perception in feature space. $\sigma (\cdot )$ is an activation function, such as $tanh(\cdot )$. Then we use an attention vector $\mathbf{a}\in {{\mathbb{R}}^{h\times 1}}$ that can be shared among the feature subgraphs, which can also be seen as a shared convolutional kernel \cite{GAT} to get the attention values ${{\omega }_{q}}={{\mathbf{a}}^{\text{T}}}\cdot \text{SLP(}\mathbf{A}_{\text{fea}}^{q}\text{)}$.

By normalizing the attention values ${{\omega }_{q}}\in {{\mathbb{R}}^{1\times n}}$, the measure $q$ based attention weight ${{\mathbf{W}}_{q}}\in {{\mathbb{R}}^{1\times n}}$ is obtained as follow: 
\begin{equation}
{{\mathbf{W}}_{q}}=softmax({{\omega }_{q}})=\frac{\exp ({{\omega }_{q}})}{\sum\nolimits_{q}{\exp ({{\omega }_{q}})}}
\end{equation}

The importance of feature subgraph $\mathbf{A}_{\text{fea}}^{q}$ is positively related to ${{\mathbf{W}}_{q}}$. In this way, by learning the attention weights ${{\mathbf{W}}_{q}}$, the importance of the measure function ${{f}_{q}}$ corresponding to each candidate feature graph $\mathbf{A}_{\text{fea}}^{q}$ is balanced. Then the final feature graph ${{\mathbf{A}}_{\text{fea}}}$ can be obtained as follows:
\begin{equation}
{{\mathbf{A}}_{\text{fea}}}=\sum\nolimits_{q}{(diag({{\mathbf{W}}_{q}})\cdot \mathbf{A}_{\text{fea}}^{q})}
\end{equation}

\subsubsection{\textbf{Embedding Layer with Local Graph Structure}}\label{sec3.1.3}
With the multi-measures fusion feature graph $\mathcal{G}_{\text{fea}}^{{}}=({{\mathbf{A}}_{\text{fea}}},\mathbf{X})$, the GNN layer can be represented as:
\begin{equation} \label{Z_fea}
\mathbf{Z}_{\text{fea}}^{i}=\sigma (\tilde{\mathbf{A}}_{\text{fea}}^{{}}\mathbf{Z}_{\text{fea}}^{i-1}\mathbf{W}_{\text{fea}}^{i}),
\end{equation}
where $\mathbf{Z}_{\text{fea}}^{i}$ is the node embedding of layer $i$ and the initial $\mathbf{Z}_{\text{fea}}^{0}=\mathbf{X}$, $\mathbf{W}_{\text{fea}}^{i}$ is the shared weight matrix of embedding layer $i$, $\tilde{\mathbf{A}}_{\text{fea}}=\tilde{\mathbf{D}}^{-\frac{1}{2}}(\mathbf{A}_{\text{fea}}+\mathbf{I}){{\tilde{\mathbf{D}}}^{-\frac{1}{2}}}$ is the normalized feature graph of $\mathbf{A}_{\text{fea}}$, $\tilde{\mathbf{D}}$ is the diagonal degree matrix of $\mathbf{A}_{\text{fea}}+\mathbf{I}$ and $\sigma(\cdot)$ is the activation function. We denote the output embedding of the last layer in the feature space as $\mathbf{Z}_{\text{fea}}$. In this way, the local structure of graph in the feature space can be preserved by obtaining the embedding $\mathbf{Z}_{\text{fea}}$.

In addition, the learned output embedding $\mathbf{Z}_{\text{ori}}$ in topology space can also be obtained by defining an embedding layer with the original adjacency matrix ${{\mathbf{A}}_{\text{ori}}}$:
\begin{equation} \label{Z_ori}
\mathbf{Z}_{\text{ori}}^{{i}}=\sigma (\tilde{\mathbf{A}}_{\text{ori}}^{{}}\mathbf{Z}_{\text{ori}}^{i-1}\mathbf{W}_{\text{ori}}^{i}).
\end{equation}
where the initial $\mathbf{Z}_{\text{fea}}^{0}=\mathbf{X}$, $\mathbf{W}_{\text{ori}}^{i}$ is the shared weight matrix of embedding layer $i$ in topology space. Hence, according to Eqs.\eqref{Z_fea} and \eqref{Z_ori}, the local structure of the graph (i.e., feature structure and original topology structure) can be preserved.

\subsection{Semantic Graph Learning for Global Structure}
Many existing GNNs use the local information of the graph to perform an 1-hop diffusion process on each layer, which may lead to misclassification due to the proximity of the nodes. In order to avoid considering only local knowledge embeddings and to sufficiently preserve the comprehensive graph structure, in addition to preserving the local graph structure defined by Eqs. \eqref{Z_fea} and \eqref{Z_ori} in Section \ref{3.1}, we further randomly sample a large number of fixed-length sentences in the graph to embed additional semantic information. Then, the frequency matrix $\mathbf{F}$ is obtained by computing the semantic similarity between pairs of nodes. Based on $\mathbf{F}$, we encode the semantic information, which is denoted as a matrix $\mathbf{P}\in {{R}^{n\times n}}$. Then, the $\mathbf{P}$-based embedding layer is defined to preserve the global graph structure.

\subsubsection{Improved calculation of PPMI matrix}
The Markov chain describing the sequence of nodes visited by a random walker is called a random walk. If the random walker is on node $\mathbf{x}_i$ at time $t$, we define the state as $s(t) = \mathbf{x}_i$. The transition probability of jumping from the current node $\mathbf{x}_i$ to one of its neighbors $\mathbf{x}_j$ is denoted as $p(s(t+1)={\mathbf{x}_{j}}\|s(t)={\mathbf{x}_{i}})$. In our problem setting, given the adjacency matrix $\mathbf{A}$, we assign:
\begin{equation}\label{RW}
p(s(t+1)={\mathbf{x}_{j}}\|s(t)={\mathbf{x}_{i}})={{{A}_{ij}}}/{\sum\nolimits_{j}{{{A}_{ij}}}}\;.
\end{equation}

To calculate frequency matrix $\mathbf{F}$ based on random walk, we define path length $q$, window size $w$ and default walks per node $\gamma$.
In addition, we divide all nodes into head and tail nodes according to their degree $deg ({{x}_{i}})$ and a related threshold $g$. For the low degree problem of tail nodes ($deg ({{x}_{i}}) \le g$), we replace unbiased sampling with degree based sampling to optimize the node sampling strategy in random walk and improve the robustness of tail nodes, so as to enhance the global structure.
Specifically, in order to achieve the above effects and reduce the computational complexity, we designed to optimize the walks per tail node as $\bar{\gamma }=\gamma \cdot deg ({{x}_{tail}})$ while the head node remains unchanged.

Next, we initialize $\mathbf{F}$ with zeros. For each ${{{x}}_{i}}$, we get path $S$ by using Eq. \eqref{RW}, uniformly sample all pairs $({{x}_{i}},{{x}_{j}})\in S$ within $w$. For each pair $({{x}_{i}},{{x}_{j}})$, frequency matrix $F_{i,j}$ and $F_{j,i}$ plus 1. Each node walks $\bar{\gamma }$ times and repeats all nodes including the head and tail nodes. The calculation process is fast because the time complexity is $O(n\gamma q^2)$. The random walk process can be conducted on different nodes simultaneously, which makes the process of calculating $\mathbf{F}$ parallel.


The value of an entry ${{F}_{i,j}}$ is the number of times that ${{\mathbf{x}}_{i}}$ occurs in context ${{\mathbf{c}}_{j}}$. Based on the frequency matrix $\mathbf{F}$, we calculate the estimated probability ${p}_{i,j}$ of the node ${{\mathbf{x}}_{i}}$ appearing in the context ${{\mathbf{c}}_{j}}$:
\begin{equation}\label{F2p}
{{p}_{i,j}}=\frac{{{F}_{i,j}}}{\sum\nolimits_{i,j}{{{F}_{i,j}}}}.
\end{equation}

After obtaining the frequency matrix $\mathbf{F}$, for each estimated probability ${p}_{i,j}$, we use the positive pointwise mutual information (PPMI) to encode semantic information in the original graph structure, which denoted as:
\begin{equation}\label{p2P}
{{P}_{i,j}}=\max \{\log (\frac{{{p}_{i,j}}}{{{p}_{i,*}}{{p}_{*,j}}}),0\},
\end{equation}
where ${{p}_{i,*}}=\sum\nolimits_{j}{{{p}_{i,j}}}$ and ${{p}_{*,j}}=\sum\nolimits_{i}{{{p}_{i,j}}}$ are the estimated probabilities of node ${{\mathbf{x}}_{i}}$ and context ${{\mathbf{c}}_{j}}$. The expectation of ${{p}_{i,j}}$ is positively related to the independence of ${{\mathbf{x}}_{i}}$ and ${{\mathbf{c}}_{j}}$.

\subsubsection{\textbf{Embedding Layer with Global Graph Structure}}
Similar to the local graph structure based embedding layer on Section \ref{sec3.1.3}, we design $\tilde{\mathbf{P}} = \mathbf{D}_{\mathbf{P}}^{-\frac{1}{2}} \mathbf{P} \mathbf{D}_{\mathbf{P}}^{-\frac{1}{2}}$ to normalize $\mathbf{P}$, where ${{\mathbf{D}}_{\mathbf{P}}}$ is the diagonal degree matrix of $\mathbf{P}$. With the normalized $\tilde{\mathbf{P}}$, the embedding layer with global graph structure can be defined as:
\begin{equation} \label{Z_sem}
\mathbf{Z}_{\mathbf{P}}^{i}=\sigma (\tilde{\mathbf{P}}\mathbf{Z}_{\mathbf{P}}^{i-1}\mathbf{W}_{\mathbf{P}}^{i}),
\end{equation}
where $\mathbf{W}_{\mathbf{P}}^{i}$ is the weight matrix, $\mathbf{Z}_{\mathbf{P}}^{i}$ is the node embedding of GNN layer $i$ in semantic space and the initial $\mathbf{Z}_{\mathbf{P}}^{0} = \mathbf{X}$. Finally, we denote the last layer of output embedding $\mathbf{Z}_{\mathbf{P}}^{i}$ in semantic space as ${{\mathbf{Z}}_{\text{sem}}}$.

\subsection{Module Aggregation}
Now there are three embeddings of which ${{\mathbf{Z}}_{\text{fea}}}$ and ${{\mathbf{Z}}_{\text{ori}}}$ are learned by local structure of graph and ${{\mathbf{Z}}_{\text{sem}}}$ is learned by global structure of graph.
To learn the optimal embedding for downstream tasks based on the three embeddings, a new module aggregation layer is adopted as follow:
\begin{equation} \label{psi_all}
{{\mathbf{Z}}_{\text{agg}}}={{\Psi }_{\text{agg}}}([{{\mathbf{Z}}_{\text{fea}}},{{\mathbf{Z}}_{\text{sem}}},{{\mathbf{Z}}_{\text{ori}}}]),
\end{equation}
where $[{{\mathbf{Z}}_{\text{fea}}},{{\mathbf{Z}}_{\text{sem}}},{{\mathbf{Z}}_{\text{ori}}}]\in {{\mathbb{R}}^{n\times {d}'\times 3}}$ is the stacked matrix of the three embeddings, ${{\Psi }_{\text{agg}}}$ denotes the attention layer with attention weight ${{\mathbf{W}}_{\text{agg}}}\in {{\mathbb{R}}^{n\times 1\times 3}}$ which represents the attention values for $n$ nodes to the three graphs.

\subsection{Objective Function}
\subsubsection{Supervised Constraint}
To make full use of the training data, we need to consider the issue of limited labeled data for semi-supervised learning. Hence, in addition to supervised learning with training data, we should derive an unsupervised regularizer for the ensemble.

For the dataset $\mathbf{X}$ and its labels $\mathbf{y}=[{{y}_{1}},\ldots, {{y}_{l}}]$, we use the aggregated embedding ${\mathbf{Z}}_{\text{agg}}$ in Eq.\eqref{psi_all} to get the prediction ${\mathbf{y}}'$ by a softmax layer. Thus, the classification loss of ML-SGNN, i.e., ${{\mathcal{L}}_{0}}$, can be calculated in the following way:
\begin{equation}\label{L0}
{\mathcal{L}}_{0} = loss({\mathbf{y}}',\mathbf{y})
\end{equation}
where $loss({\mathbf{y}}',\mathbf{y})$ calculates the difference between prediction $y_{i}^{'}$ and the real label ${{y}_{i}}$ as follows:
\begin{equation}\label{cross_entropy}
loss({\mathbf{y}}',\mathbf{y})=- \sum_{i=1}^C {{y}_{i}} \ln  y_{i}^{'}
\end{equation}

\subsubsection{Unsupervised Constraint}\label{sec3.4.2}
In addition to using training data for supervised learning, we derive an unsupervised regularizer. Considering that the graph embedding learning method is prone to overfitting, we introduce an unsupervised constraint on embeddings to improve generalization.

Specifically, to minimize the difference among the three embeddings ${{\mathbf{Z}}_{\text{fea}}}$,  ${{\mathbf{Z}}_{\text{ori}}}$ and ${{\mathbf{Z}}_{\text{sem}}}$, we adopt a sparse-inducing ${{l}_{2,1}}$ regularizer constraint \cite{nie2010flexible} on ${{\mathbf{Z}}_{\text{fea}}}$, ${{\mathbf{Z}}_{\text{ori}}}$: 
\begin{equation}\label{La}
{{\mathcal{L}}_{reg_a}}=\|{{\mathbf{Z}}_{\text{fea}}}-{{\mathbf{Z}}_{\text{ori}}}{\|_{2,1}},
\end{equation}
and on ${{\mathbf{Z}}_{\text{sem}}}$, ${{\mathbf{Z}}_{\text{ori}}}$:
\begin{equation}\label{Lb}
{{\mathcal{L}}_{reg_b}}=\|{{\mathbf{Z}}_{\text{sem}}}-{{\mathbf{Z}}_{\text{ori}}}{\|_{2,1}}.
\end{equation}

By minimizing the loss function ${{\mathcal{L}}_{reg_a}}$ and ${{\mathcal{L}}_{reg_b}}$, the embeddings obtained by each module should then be similar to a large extent. From another point of view, we can also regard the loss functions as training ${{\mathbf{Z}}_{\text{fea}}}$ and ${{\mathbf{Z}}_{\text{sem}}}$ against ${{\mathbf{Z}}_{\text{ori}}}$.

\subsubsection{Optimization Objective}
We utilize node features $\mathbf{X}\in {{\mathbb{R}}^{n\times d}}$ and original graph $\mathbf{A}\in {{\mathbb{R}}^{n\times n}}$ to construct the topology graph, feature graph and semantic graph respectively, and construct corresponding graph convolution modules, which are aggregated by attention. Combining the loss functions Eqs.\eqref{L0}, \eqref{La} and \eqref{Lb}, we have the following overall objective function:
\begin{equation} \label{loss_function}
\mathcal{L}={{\mathcal{L}}_{0}}+{{\alpha }}{{\mathcal{L}}_{reg_a}}+{{\beta }}{{\mathcal{L}}_{reg_b}},
\end{equation}
where $\alpha $ and $\beta $ are parameters to control the strength of unsupervised constraints. With the guide of labeled data, the \textbf{Adam} \cite{Adam} optimizer is utilized to optimize the proposed model.

\begin{table}[]
\centering
\caption{Overview of the Six Datasets}
\begin{tabular}{@{}ccccc@{}}
\toprule
Dataset     & Nodes & Features & Edges  & Classes \\ \midrule
Citeseer    & 3327  & 3703     & 4732   & 6       \\
UAI2010     & 3067  & 4973     & 28311  & 19      \\
ACM         & 3025  & 1870     & 13128  & 3       \\
BlogCatalog & 5196  & 8189     & 171743 & 6       \\
Flickr      & 7575  & 12047    & 239738 & 9       \\
CoraFull    & 19793 & 8710     & 65311  & 70      \\ \bottomrule
\end{tabular}
\label{datasets}
\end{table}

\section{Experiments}\label{sec4}
\subsection{Datasets}
We evaluated our proposed method on six common real-world datasets, including Citeseer \cite{giles_citeseer_1998}, UAI2010 \cite{AMGCN}, ACM \cite{yun2019graph}, CoraFull \cite{bojchevski2018deep}, BlogCatalog \cite{meng2019co} and Flickr \cite{demonet}. The first four are citation networks, and the last two are social relationship network and graphic social network respectively. The statistics of these datasets are shown in Table \ref{datasets}:

\begin{table*}[]
\centering
\setlength{\tabcolsep}{2.5mm}
\caption{Node classification performance(\%). The best and the runner-up are shown in boldface and underline.}
\begin{tabular}{cccccccccc}
\hline
\multirow{2}*{Method} & \multicolumn{3}{c}{Citeseer} & \multicolumn{3}{c}{UAI2010} & \multicolumn{3}{c}{ACM} \\ \cline{2-10}
&20L/C&40L/C&60L/C&20L/C&40L/C&60L/C&20L/C&40L/C&60L/C 
\\ \hline

DeepWalk   & 43.47 & 45.15 & 48.86 & 42.02 & 51.26 & 54.37 & 62.69 & 62.90    & 67.03 \\
GCN       & 70.30  & 73.10  & 74.48 & 49.88 & 51.80  & 54.40  & 87.80  & 89.06 & 90.54 \\

GAT       & 72.50  & 73.04 & 74.76 & 56.92 & 63.74 & 68.44 & 87.36 & 88.60  & 90.40  \\
DEMO-Net   & 69.50  & 70.44 & 71.86 & 23.45 & 30.29 & 34.11 & 84.48 & 85.70  & 86.55 \\
MixHop    & 71.40  & 71.48 & 72.16 & 61.56 & 65.05 & 67.66 & 81.08 & 82.34 & 83.09 \\
DGCN      & 71.30  & 73.40  & \underline{76.90}  & 58.29 & 64.12 & 68.55 & 87.40  & 88.10  & 90.60  \\
AM-GCN    & 73.10  & \underline{74.70}  & 75.56 & \underline{70.10}  & \underline{73.14} & 74.40  & \underline{90.40}  & \underline{90.76} & \underline{91.42} \\
Tail-GNN  & \underline{73.30}  & 73.60  & 74.20  & 69.42 & 72.24 & \underline{75.05} & 89.80  & 90.30  & 90.60  \\
ML-SGNN&\textbf{74.20}&\textbf{76.20}&\textbf{77.50}&\textbf{73.90}&\textbf{76.60}&\textbf{77.90}&\textbf{91.00}&\textbf{91.20}&\textbf{91.50}  \\ \hline
\end{tabular}

\begin{tabular}{cccccccccc}
\hline
\multirow{2}*{Method} & \multicolumn{3}{c}{BlogCatalog} & \multicolumn{3}{c}{Flickr} & \multicolumn{3}{c}{CoraFull} \\ \cline{2-10}
&20L/C&40L/C&60L/C&20L/C&40L/C&60L/C&20L/C&40L/C&60L/C 
\\ \hline
DeepWalk   & 38.67 & 50.80 & 55.02 & 24.33 & 28.79 & 30.10 & 29.33 & 36.23 & 40.62 \\
GCN        & 69.84 & 71.28 & 72.66 & 41.42 & 45.48 & 47.96 & 56.68 & 60.60 & 62.04 \\
GAT        & 64.08 & 67.40 & 69.95 & 38.52 & 38.44 & 38.96 & 58.44 & 62.98 & 64.38 \\
DEMO-Net   & 54.19 & 63.47 & 76.81 & 34.89 & 46.57 & 57.30 & 54.50 & 60.28 & 61.58 \\
MixHop     & 65.46 & 71.66 & 77.44 & 39.56 & 55.19 & 64.96 & 47.74 & 57.20 & 60.18 \\
DGCN       & 73.82 & 76.47 & 78.12 & 69.78 & 73.56 & 76.06 & 57.16 & \underline{64.12} & \underline{65.83} \\
AM-GCN     & 81.98 & 84.94 & 87.30  & \underline{75.26} & \underline{80.06} & \underline{82.10} & \underline{58.90}  & 63.62 & 65.36 \\
Tail-GNN  & \underline{83.46}  & \underline{86.21}  & \underline{88.31}  & 59.34 & 63.88 & 67.78 & 57.25  & 59.31  & 60.41  \\
ML-SGNN&\textbf{87.35}&\textbf{88.96}&\textbf{89.92}&\textbf{77.14}&\textbf{81.31}&\textbf{82.75}&\textbf{60.82}&\textbf{64.39}&\textbf{66.32}  \\ \hline	 
\end{tabular}

\label{ACC}
\end{table*}

\begin{itemize}
\item\textbf{{Citeseer}}\cite{GCN}: 
This dataset is a citation network of research papers, which are divided into six categories. The citation network takes 3327 scientific papers as nodes and 4732 citation links as edges. The feature of each node in the dataset is a word vector to describe whether the paper has the corresponding words or not.
\item\textbf{{CoraFull}}\cite{bojchevski2018deep}:	
Similar to the Citeseer dataset, CoraFull is a well-known citation network labeled based on the paper topic which contains 19793 scientific publications. CoraFull is classified into one of 70 categories, where nodes represent papers and the edges represent citations. 
\item\textbf{{UAI2010}}\cite{AMGCN}:
This dataset contains 3067 nodes and 28311 edges which has been tested for node classification in \cite{AMGCN}.
\item\textbf{{ACM}}\cite{yun2019graph}:
It is a citation network dataset, where nodes represent papers and node features are constructed by the keywords. Papers are divided into 3 categories according to their types of conferences. 
\item\textbf{{BlogCatalog}}\cite{meng2019co}:
This dataset is a social relationship network. The graph is composed of bloggers and their social relationships (such as friends). Node attributes are constructed by keywords in user profile. The labels represent bloggers' interests. All nodes are divided into six categories.
\item\textbf{{Flickr}}\cite{demonet}:
It is a graphic social network, where nodes represent users and edges correspond to the friendships among users. All the nodes are divided into 9 classes according to the interest groups of users.

\end{itemize}

\subsection{Baselines}
We compare ML-SGNN with eight state-of-the-art methods,
including six GNN-based methods, i.e., GCN \cite{GCN},
GAT \cite{GAT}, AM-GCN \cite{AMGCN},
DEMO-Net \cite{demonet}, MixHop \cite{Mixhop}, Tail-GNN \cite{tail_gnn}
and two semantic-based methods, i.e., Deepwalk \cite{deepwalk}  and DGCN \cite{DGCN}. We provide the corresponding code websites for all methods.

\begin{itemize}
\item\textbf{{Deepwalk}}\footnote[1]{https://github.com/phanein/deepwalk/}\cite{deepwalk} 
is one of the most distinguished methods in network representation learning, which is basically to map the relationship and structure properties of the nodes in the graph to a new vector space. 

\item\textbf{{GCN}}\footnote[2]{https://github.com/tkipf/pygcn/}\cite{GCN} is a state-of-the-art model which learns node representations by aggregating information from neighbors.

\item\textbf{{GAT}}\footnote[3]{https://github.com/PetarV-/GAT/}\cite{GAT} is a GCN model using attention mechanism to aggregate all neighbor information. 

\item\textbf{{DEMO-Net}}\footnote[4]{https://github.com/jwu4sml/DEMO-Net/}\cite{demonet} is a generic degree-specific GNN-based method that recursively identifies 1-hop neighborhood structures.

\item\textbf{{MixHop}}\footnote[5]{https://github.com/samihaija/mixhop/}\cite{Mixhop} is a GNN-based method which mixes information of neighbors from different orders and learns node representation. 

\item\textbf{{DGCN}}\footnote[6]{https://github.com/ZhuangCY/DGCN/}\cite{DGCN} is a GNN-based method that considers the global consistency and local consistency in semi-supervised learning. 

\item\textbf{{AM-GCN}}\footnote[7]{https://github.com/zhumeiqiBUPT/AM-GCN/}\cite{AMGCN} enhances the fusion capability of GNN for classification by extracting more correlation information from both node features and topological structures nicely. 

\item\textbf{{Tail-GNN}}\footnote[8]{https://github.com/shuaiOKshuai/Tail-GNN}\cite{tail_gnn} is a GNN-based method that can improve tail node embeddings during neighborhood aggregation.
\end{itemize}

\subsection{Experimental Settings}
For our method, we build the training set, validation set and test set by following the setting of \cite{AMGCN}. Specifically, in order to fairly compare the effects of different methods, we create three levels of training set that each class contains 20, 40 and 60 labeled nodes. 1000 nodes are chosen as the test set. All baselines are initialized with the same parameters suggested by their papers and we also further carefully turn parameters for optimal performance. Accuracy (ACC) is used to evaluate performance of the models.

To allow a fair comparison between ML-SGNN and DGCN, we followed the parameter settings of DGCN when creating the global graph structure, including the negative sampling size of 2, which is to calculate each entry of $\mathbf{P}$; default walks per node $\gamma$ of 100; path length $q$ of 3, although in \cite{DGCN} these parameters were confirmed to have little effect on final performance.

In addition, we set the hidden layer dimension of graph convolution to 512 or 768 and the output dimension to 32 or 128 or 256.
Considering the problem of over smoothing caused by multi-layer GNN \cite{li2018deeper}, the number of graph convolution layers is set to 2 simultaneously. 
The learning rate used in the Adam optimizer ranges from $1e - 4$ to $5e - 4$,
the weight decay $\varpi $ is set from $5e - 3$ to $5e - 4$,
and the dropout rate is set to 0.5.
Besides,
the activation function we used in Eqs.\eqref{Z_fea}, \eqref{Z_ori} and \eqref{Z_sem} is $ReLU(\cdot)$,
the constraint weight parameter $\alpha $ and $\beta $ in Eq.\eqref{loss_function} are set from $1e - 3$ to $1e + 3$.
All experiments are run for three times, and the best performance is reported.

\subsection{Performance Comparisons and Discussions}\label{4.4}
The node classification results are shown in Table \ref{ACC}, where L/C is the number of labeled nodes of each type. It can be observed that:
\begin{itemize}
\item	Compared to all baselines, ML-SGNN performs best on all datasets with all label rates. In particular, compared to the runner-up, ML-SGNN achieves the improvement of ACC by 3.89\% on BlogCatalog with 20L/C and 3.8\% on UAI2010 with 20L/C. The results demonstrate the effectiveness of ML-SGNN, and show that our method can extract more useful feature information than the single measure based method.

\item	Compared with the advantages of semantic-based DGCN, the performance improvement of ML-SGNN is more prominent. Particularly, on the highly sparse datasets CoraFull, our method outperforms the best baseline by 1.92\%. It further shows the advantage of semantic context coding for global graph structure preservation.

\item	Since multi-measure can extract more useful feature information than traditional way, the improvement is more obvious on datasets with rich feature information, especially UAI2010, BlogCatalog and Flickr. It implies that ML-SGNN provides a more optimal and suitable feature graph to the optimizer for learning the node representation.

\end{itemize}

\begin{figure}
\centering
\begin{subfigure}{6cm}
\centering
\includegraphics[width=0.8\textwidth]{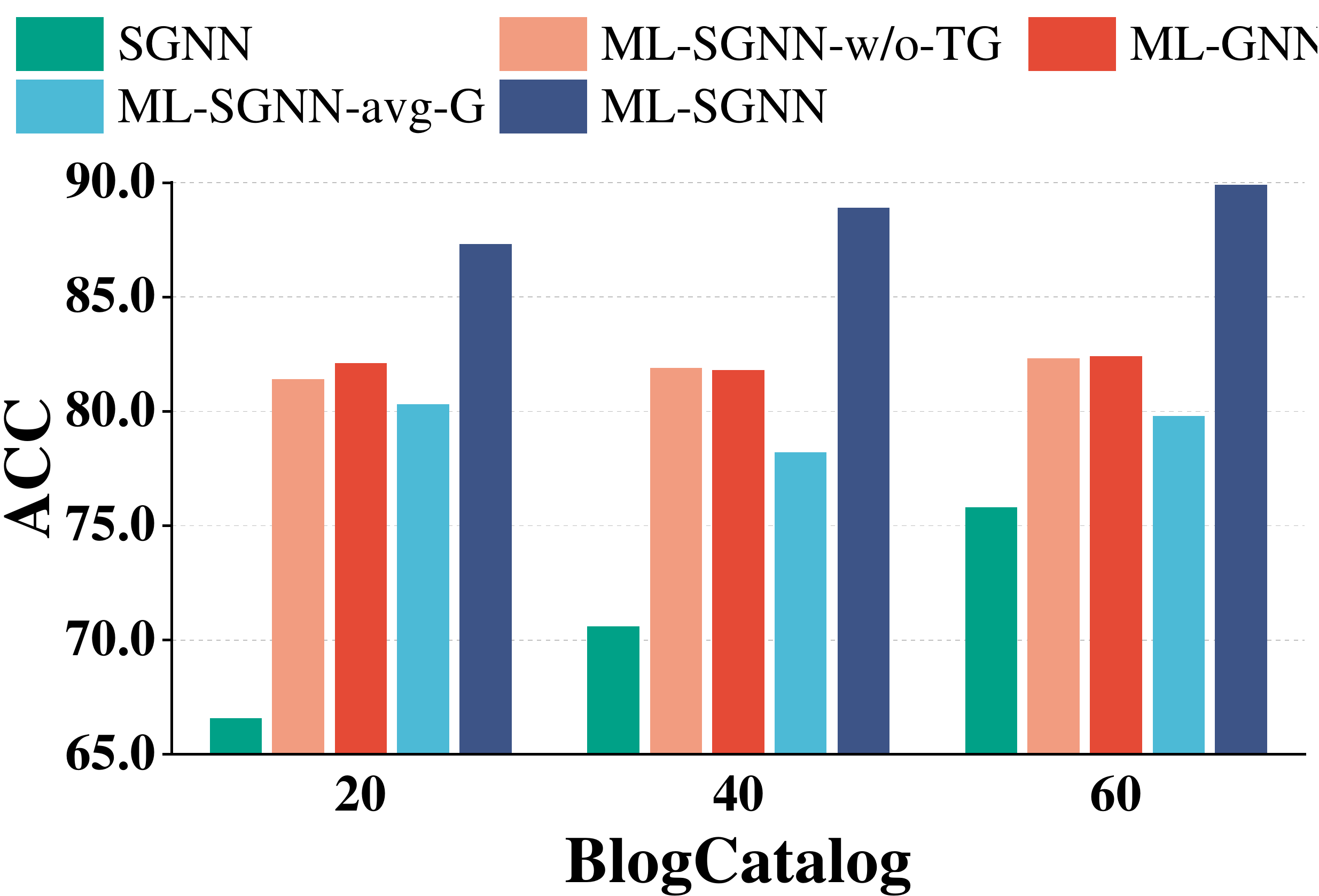}
\end{subfigure}
\begin{subfigure}{6cm}
\centering
\includegraphics[width=0.8\textwidth]{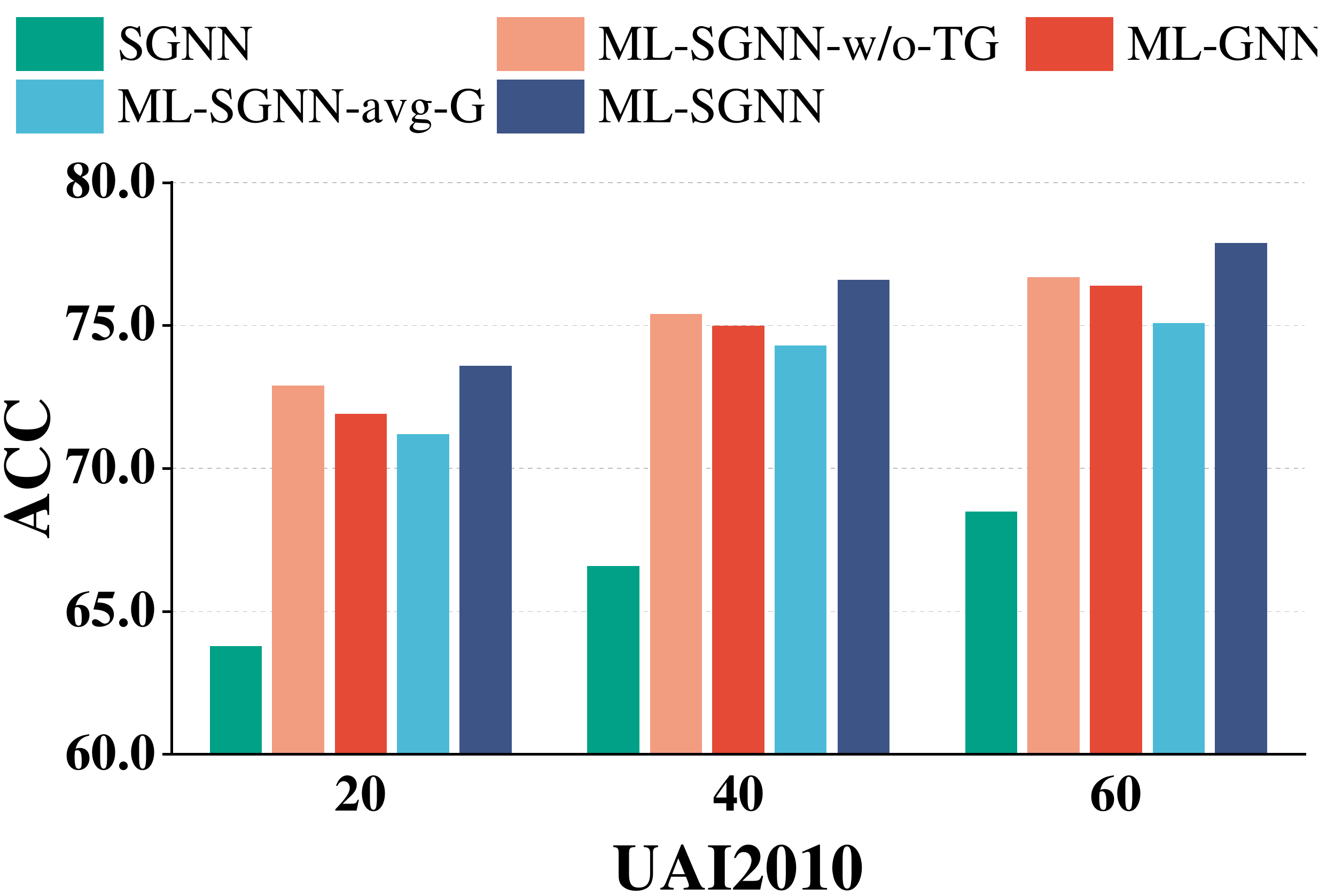}
\end{subfigure}
\caption{Performance of ML-SGNN and its variants.}
\label{Ablation1}
\end{figure}
\begin{figure}
\centering
\begin{subfigure}{6cm}
\centering
\includegraphics[width=0.8\textwidth]{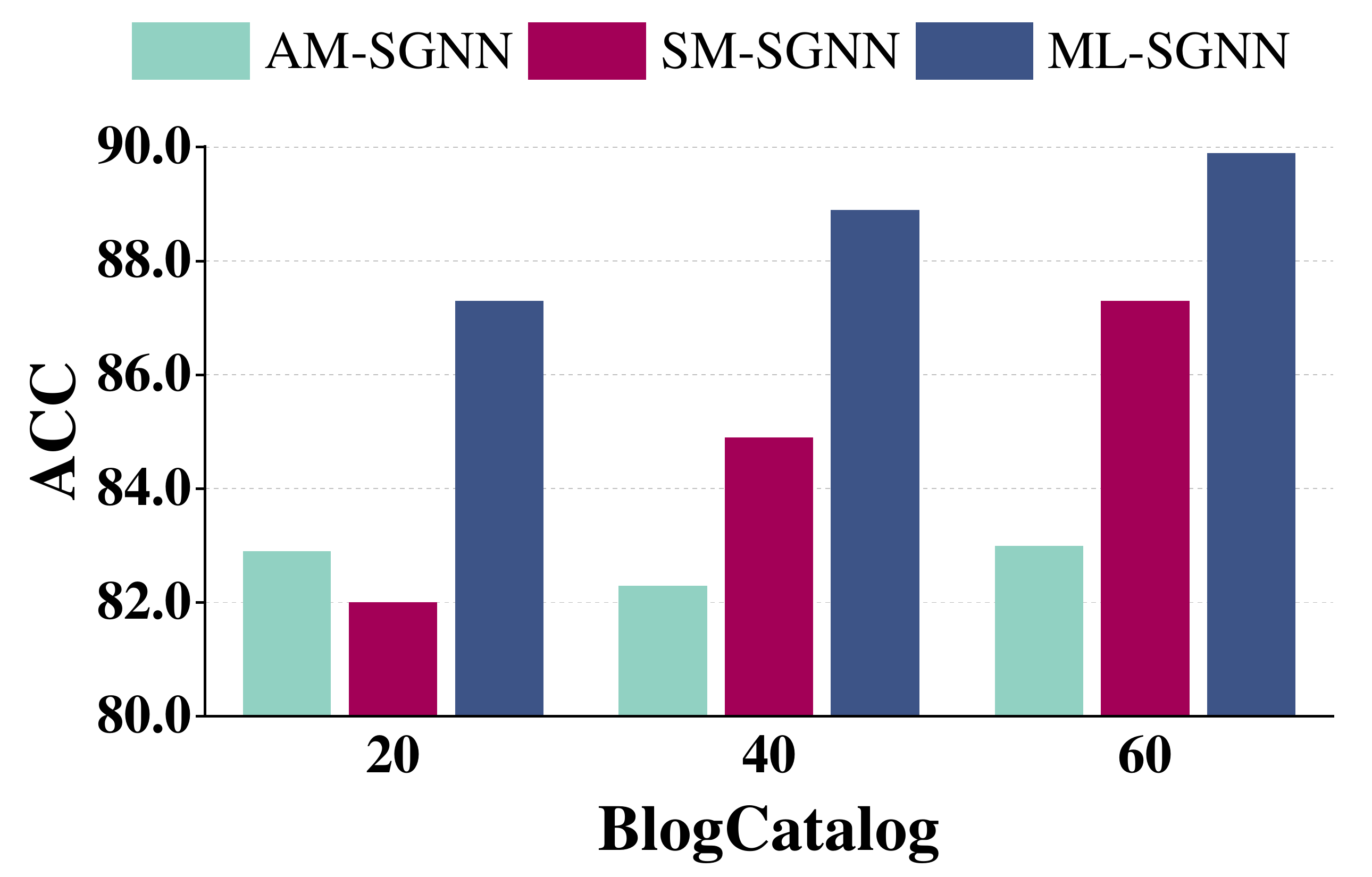}
\end{subfigure}
\begin{subfigure}{6cm}
\centering
\includegraphics[width=0.8\textwidth]{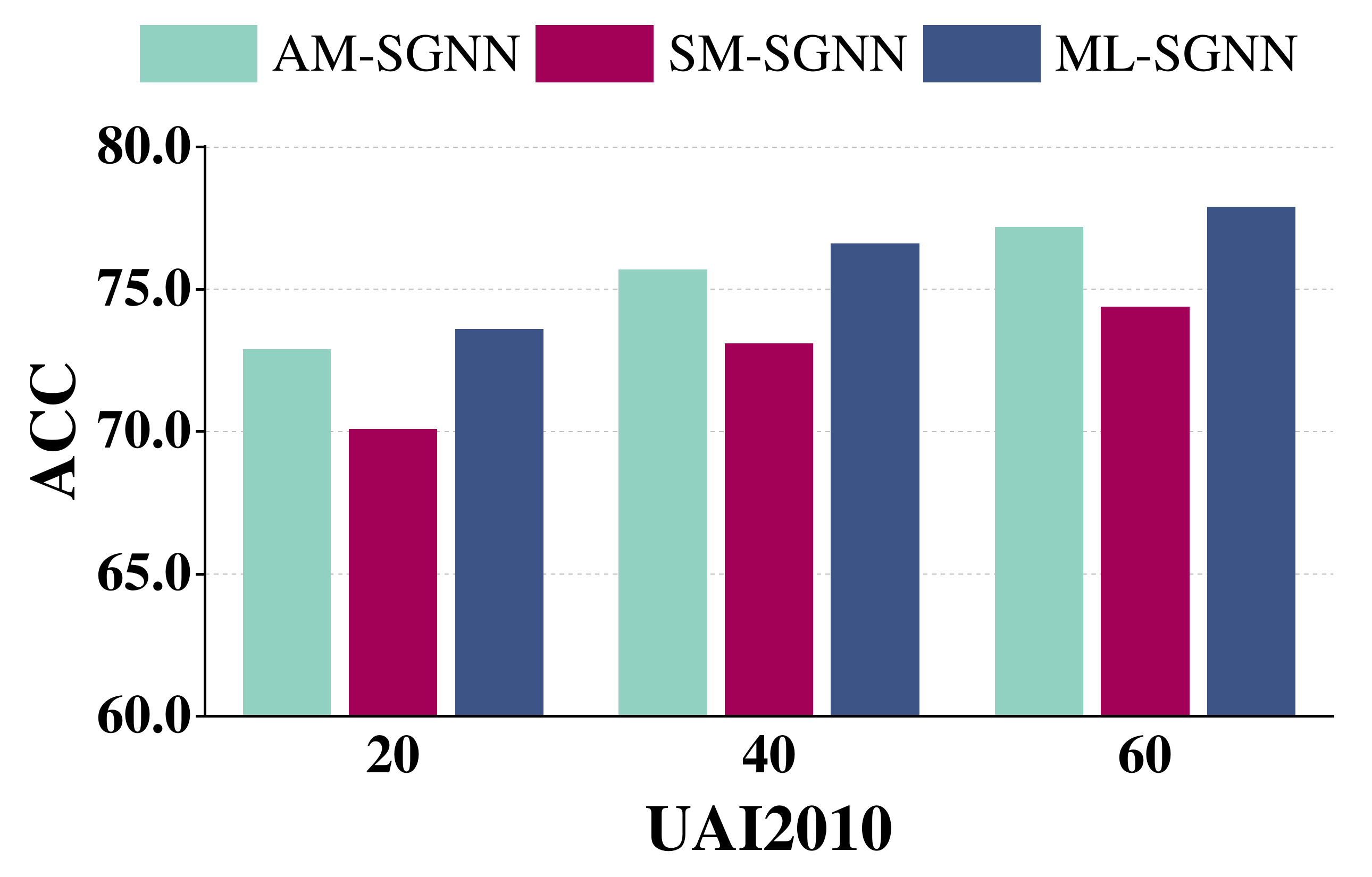}
\end{subfigure}
\caption{Comparision of ML-SGNN with average-measure and single-measure.}
\label{Ablation2}
\end{figure}

\subsection{Statistical Tests of Performance}\label{Statistical_tests}
\begin{figure}
\centering
\includegraphics[width=0.7\linewidth]{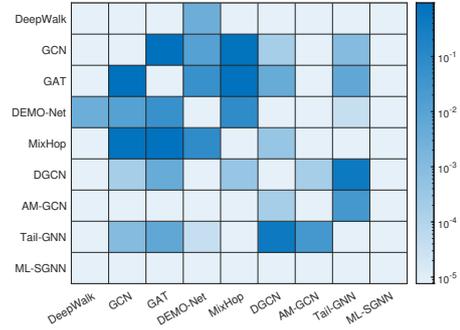}
\caption{Heatmap of p-values obtained from Wilcoxon test.}
\label{fig:heatmap}
\end{figure}

In order to explore whether there is significant difference in classification accuracy between our method and baselines, we perform the Wilcoxon signed rank test on the classification accuracy of 9 methods. The null hypothesis of this test is that the median accuracy of the two methods is equal. If the p-value of the test is less than 0.05, the null hypothesis can be rejected, and we can conclude that the median ranking of accuracy of the two models on all datasets is not equal, that is, there is a significant difference in the classification performance of the two methods. The heatmap result of the p-values is shown in Fig. \ref{fig:heatmap}. It shows that our ML-SGNN significantly outperforms all baselines at p < 0.05, which can prove to be a competitive GNN method for semi-supervised classification tasks.

\begin{figure}
\centering
\includegraphics[width=0.7\linewidth]{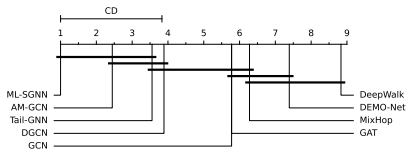}
\caption{Critical Difference (CD) diagram from Friedman test.}
\label{fig:plteps}
\end{figure}

To further analyze the performance of these methods, we additionally perform the Nemenyi non-parametric statistical test and plot the critical difference diagram. The null hypothesis is that the average rank of 9 classifiers on all datasets is the same. The results are presented in Fig. \ref{fig:plteps}. It suggests that our ML-SGNN significantly superior to DGCN, GCN, GAT, MixHop, DEMO-net and DeepWalk based on statistics. Although the performance of Tail-GNN, AM-GCN and our ML-SGNN are not statistically different, the ranking value of our method is obviously superior to the other two baseline methods.

\subsection{Ablation Study}\label{4.5}
ML-SGNN adaptively aggregates embeddings of three candidate graphs, denoted as fusion feature graph, original topological graph, and semantic graph, to generate optimal graph embedding. In order to verify the effectiveness of different components of ML-SGNN, we devise six variants and compare their classification performance with ML-SGNN on datasets BlogCatalog and UAI2010. Specifically, we design SGNN, ML-SGNN-w/o-TG, ML-GNN by removing feature graph, topology graph and semantic graph respectively. ML-SGNN-avg-G is designed by replacing the candidate graph aggregation layer with the average aggregation layer.
The replacement of the multi-measure attention layer with a single measure layer, where the feature graph is obtained by cosine measure, is denoted as AM-SGNN.
SM-SGNN refers to the replacement of the multi-measure attention layer with an average aggregation layer, i.e. the feature graph is obtained by averaging the three subgraphs.
The results of ACC are shown in Fig. \ref{Ablation1} and Fig. \ref{Ablation2}.



\begin{figure*} 
\centering
\begin{subfigure}{5cm}
\centering
\includegraphics[width=1\textwidth]{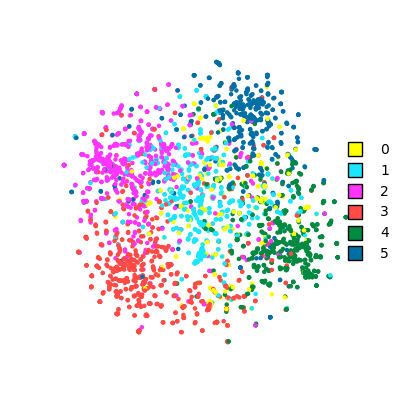}
\subcaption{GCN}
\end{subfigure}
\begin{subfigure}{5cm}
\centering
\includegraphics[width=1\textwidth]{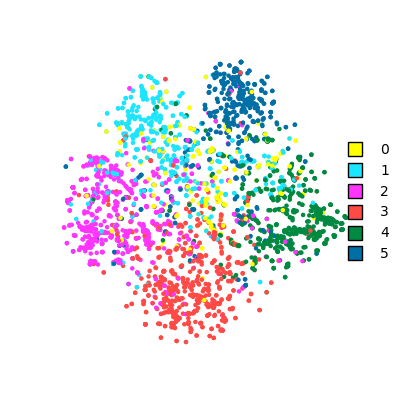}
\subcaption{AMGCN}
\end{subfigure}
\begin{subfigure}{5cm}
\centering
\includegraphics[width=1\textwidth]{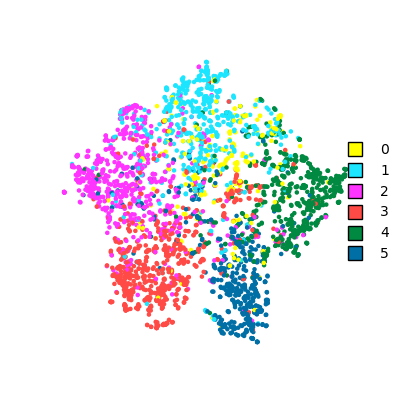}
\subcaption{ML-SGNN}
\end{subfigure}
\caption{The t-SNE visualization of node embeddings on Citeseer.}
\label{Visualization}
\end{figure*}

\subsubsection{Effectiveness of Components}\label{components}

SGNN, ML-SGNN-w/o-TG, ML-GNN are chosen to verify the impact of the three graphs. From Fig. \ref{Ablation1}, we can observe that:

\begin{itemize}
\item On all datasets with all label rates, compared with ML-SGNN, the performance of ML-GNN declines at different extent. The results show that the semantic graph is effectively capable of preserving the global structure, whereas the other two graphs fail to, which reflects the necessity of semantic graph aggregation in GNN.
\item Compared with DGCN, SGNN can be regarded as a model that improves the limited neighbors of the tail node to preserve the global graph structure. Although there is no multi-measure based feature fusion, which makes it inferior to the proposed ML-SGNN, it ensures the robustness of the tail nodes for better preservation of the global graph structure than DGCN since it considers the optimization of the tail nodes. A comparison with Table \ref{ACC} and Fig. \ref{Ablation1} shows that it is superior to DGCN in most cases.

\item The decline of performance of these three variants varies on different datasets, which shows that all these components are necessary to take into consideration. Attention mechanism plays an significant role in balancing the importance of these candidate graphs. 
\end{itemize}

\subsubsection{Effectiveness of Attention Learning}
ML-SGNN-avg-G is designed to evaluate whether ML-SGNN can effectively learn the importance of candidate graphs by attention mechanism. From the observation of Fig. \ref{Ablation1}, the following conclusions can be drawn:
\begin{itemize}
\item The performance of ML-SGNN is considerably better than ML-SGNN-avg-G, which shows that weight learning through the attention layer is effective. 
\item The performance of ML-SGNN-avg-G on BlogCatalog is significantly declined, because the distributions of attention values are relatively scattered on these datasets. However, ML-SGNN-avg-G averages three candidate graphs simply, which compromises of the influence of the node features and reduces the performance.
\item Therefore, ML-SGNN is able to adaptively adjust each graph to obtain the best weight, which can avoid the negative impact of data characteristics.
\end{itemize}

\subsubsection{Effectiveness of Multi-Measure}
\begin{figure*} 
\centering
\begin{subfigure}{5cm}
\centering
\includegraphics[width=1\textwidth]{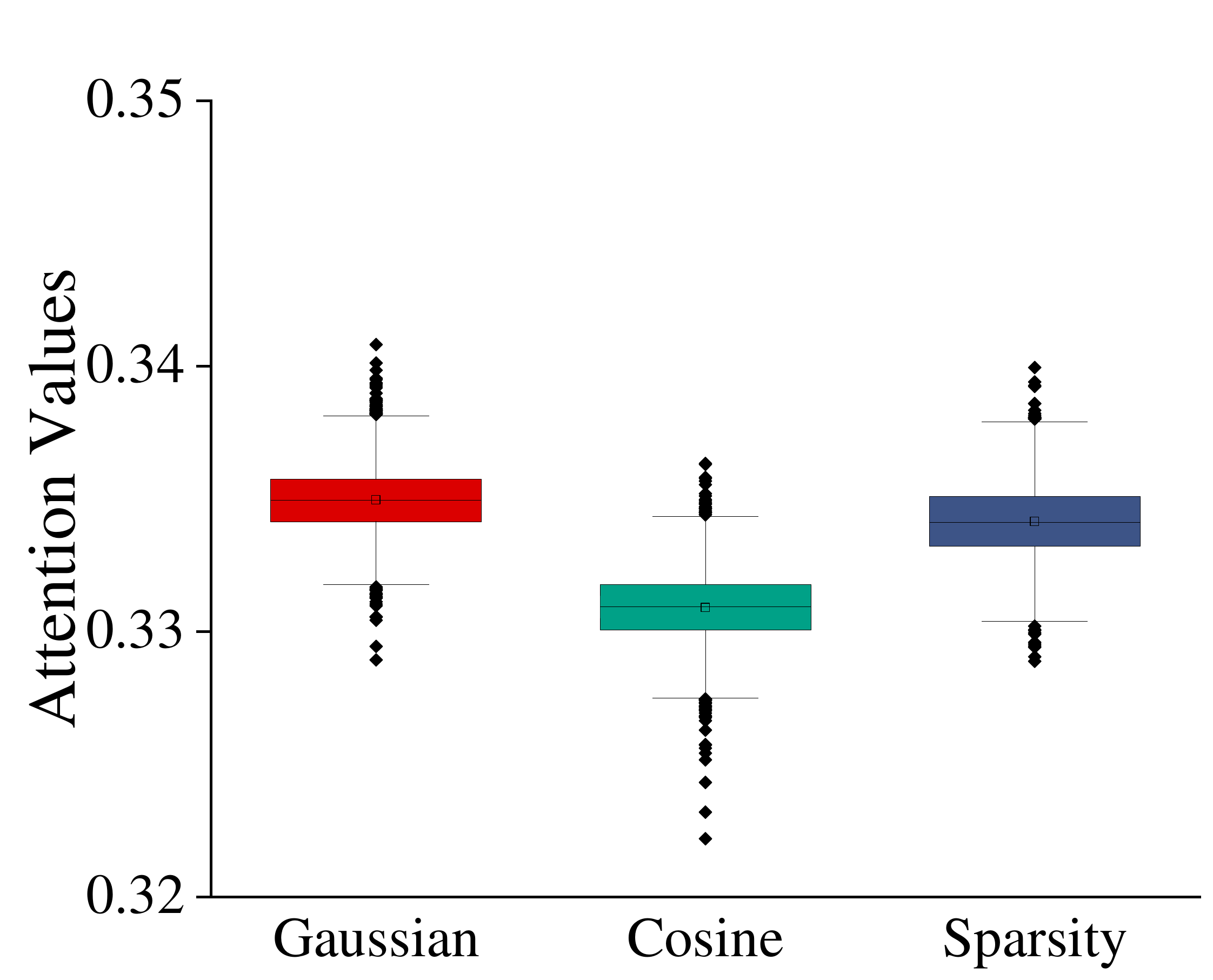}
\subcaption{Citeseer}
\end{subfigure}
\begin{subfigure}{5cm}
\centering
\includegraphics[width=1\textwidth]{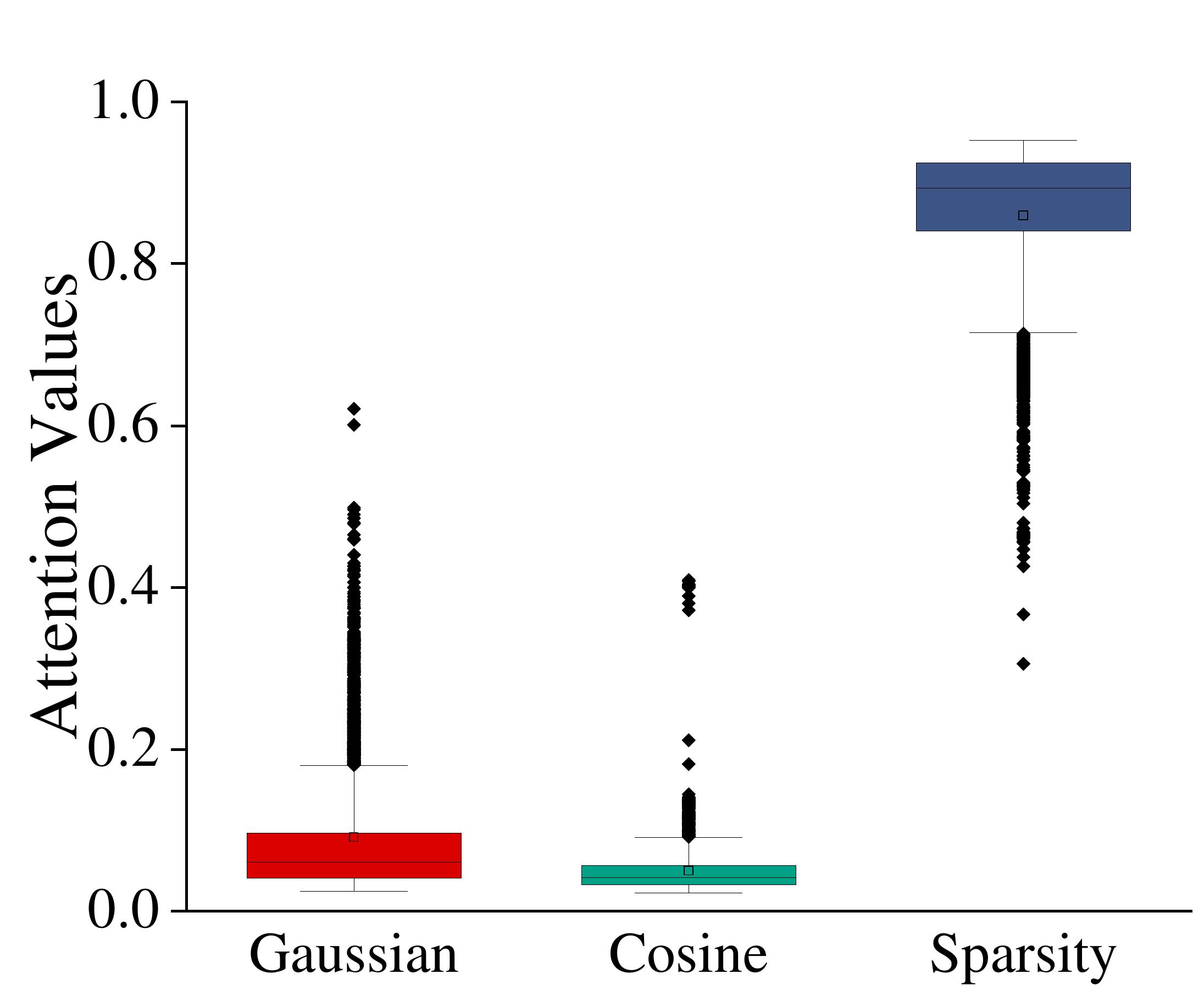}
\subcaption{UAI2010}
\end{subfigure}
\begin{subfigure}{5cm}
\centering
\includegraphics[width=1\textwidth]{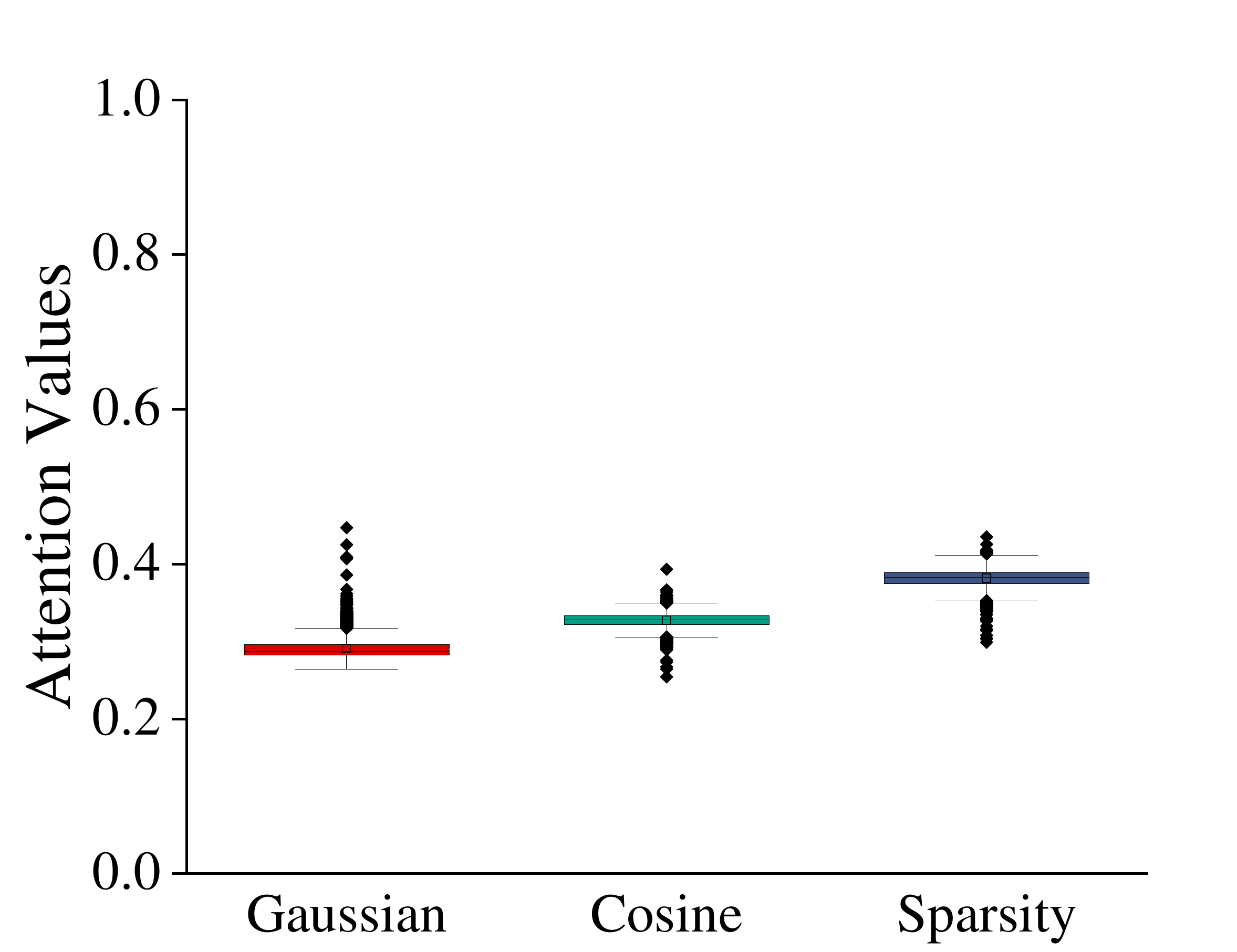}
\subcaption{ACM}
\end{subfigure}
\begin{subfigure}{5cm}
\centering
\includegraphics[width=1\textwidth]{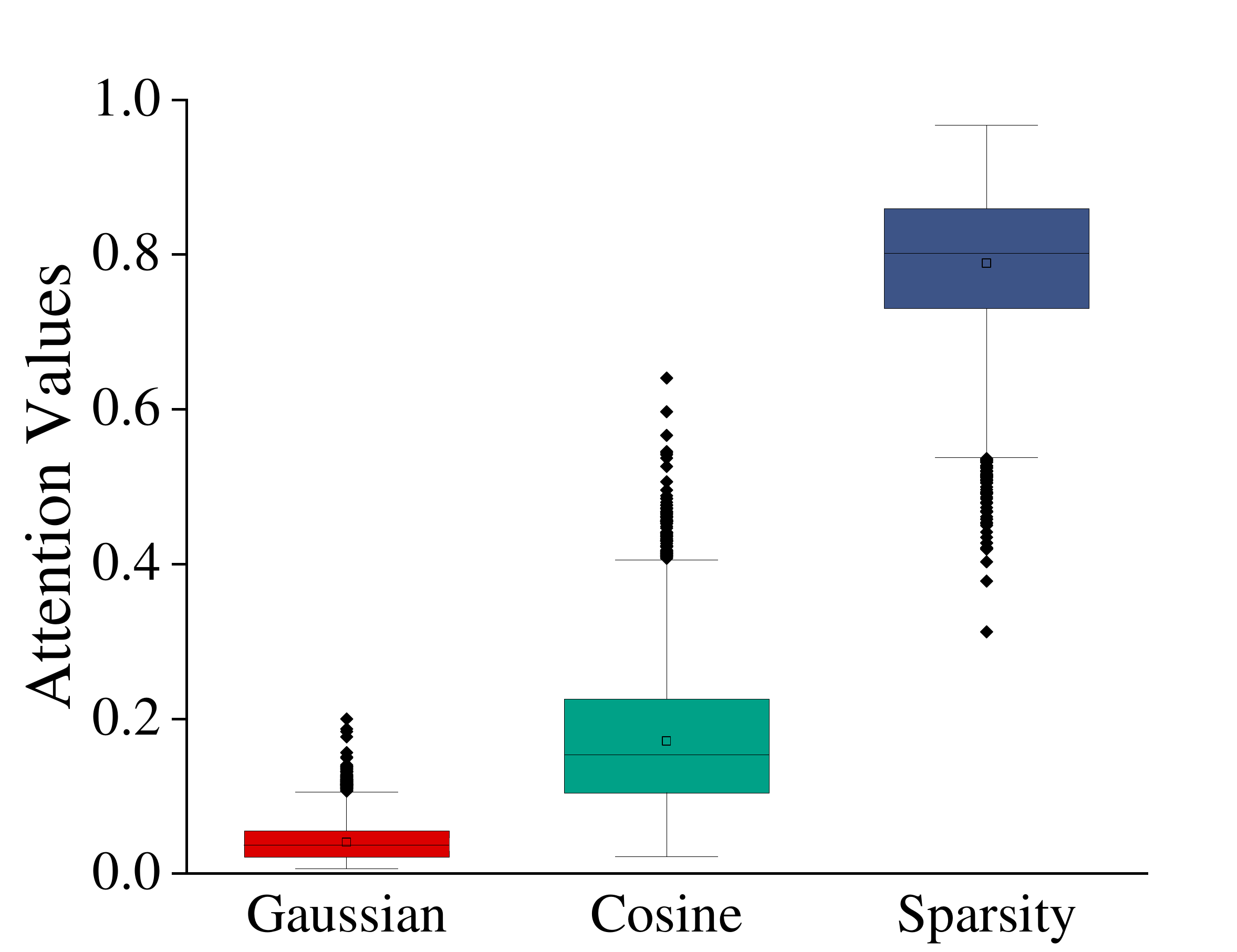}
\subcaption{BlogCatalog}
\end{subfigure}
\begin{subfigure}{5cm}
\centering
\includegraphics[width=1\textwidth]{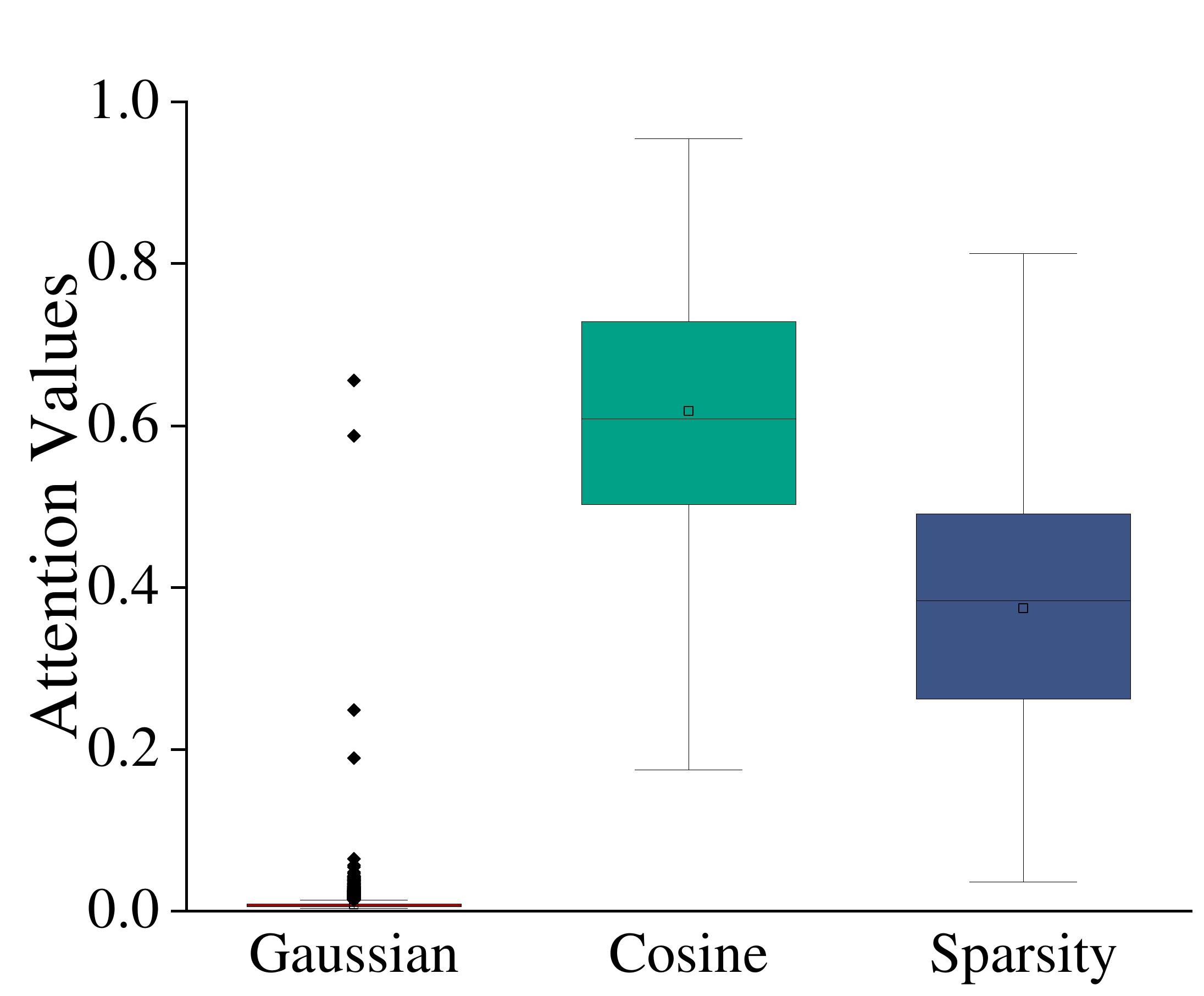}
\subcaption{Flickr}
\end{subfigure}
\begin{subfigure}{5cm}
\centering
\includegraphics[width=1\textwidth]{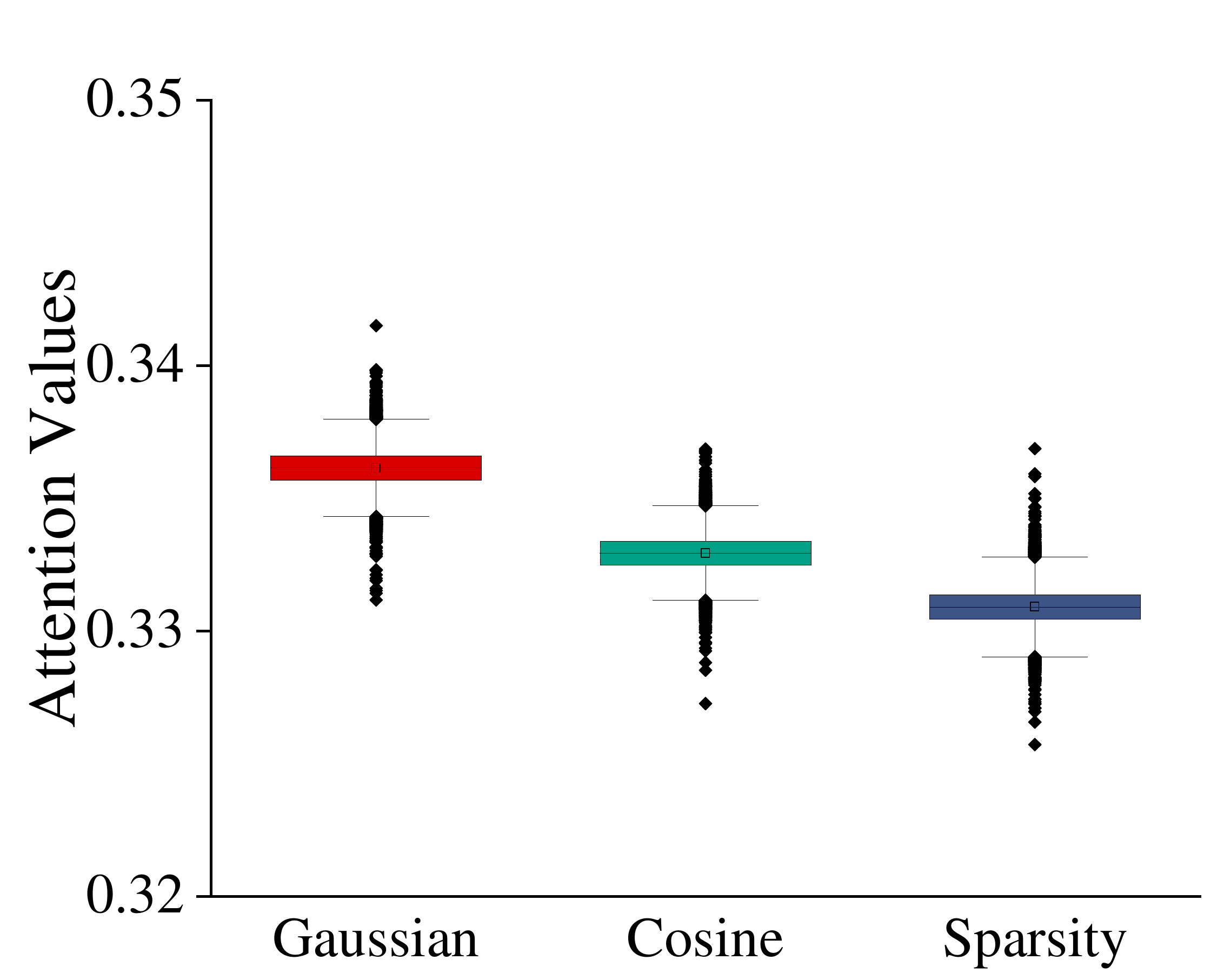}
\subcaption{CoraFull}
\end{subfigure}
\caption{Boxplots illustration of attention distribution of multi-measures on six datasets.}
\label{Boxplot_mea}
\end{figure*}

AM-SGNN and SM-SGNN are used to test the effectiveness of multi-measure and attention mechanism in ML-SGNN, which is shown in Fig. \ref{Ablation2}. Since multi-measures can consider more individual perspectives and extract more information than single-measure, AM-SGNN has better effects than SM-SGNN on BolgCatalog with 20 label rate and on UAI2010 with all label rates. ML-SGNN has the best effect on all datasets. Besides, compared with Fig. \ref{Ablation2} and Fig. \ref{Boxplot_mea}, ML-SGNN performs better than AM-SGNN, and the improvement is related to the sensitivity of data to each measure, which is due to the attention mechanism of adaptive learning.

\subsection{Statistical analysis of attention value}\label{4.6}
In our experiments, three different measure functions are utilized in Eq.\eqref{psi_measures}. In order to verify whether ML-SGNN can adaptively fuse the measures through trade-offs, we analyze the distribution of attention values in ${\Psi }_{\text{fea}}$ on all datasets with 20 label rate, as shown in Fig. \ref{Boxplot_mea}.

It can be seen from the boxplots that the data distribution of all attention values is relatively concentrated. In addition, there is almost no case where the attention value of any measure which is much larger than that of the other two. Specifically, Sparsity method has the highest adaptability on UAI2010, ACM and BlogCatalog, and Gaussian method has the highest adaptability on Citeseer and CoraFull. It is worth noting that the commonly used Cosine method has less weight than others outside the dataset Flickr. The above analysis shows that the ML-SGNN method can adaptively fuse appropriate measures from different datasets.



\subsection{Visualization}\label{tsne}
In order to more visually show the effectiveness of our proposed model, we perform a visualization task on Citeseer dataset, which is shown in Fig. \ref{Visualization}. We plot the output embeddings of the last layer in the test set with t-SNE \cite{t_SNE}, which are colored by real labels. Apparently, compared to the baselines, the learned embeddings of ML-SGNN has relatively the highest intra-class similarity and the best discrimination boundaries among different classes, which further demonstrates the advantages of our model.

\subsection{\textbf{Parameter Sensitivity}}\label{para}
\begin{figure}
\centering
\begin{subfigure}{0.23\textwidth}
\includegraphics[width=1\textwidth]{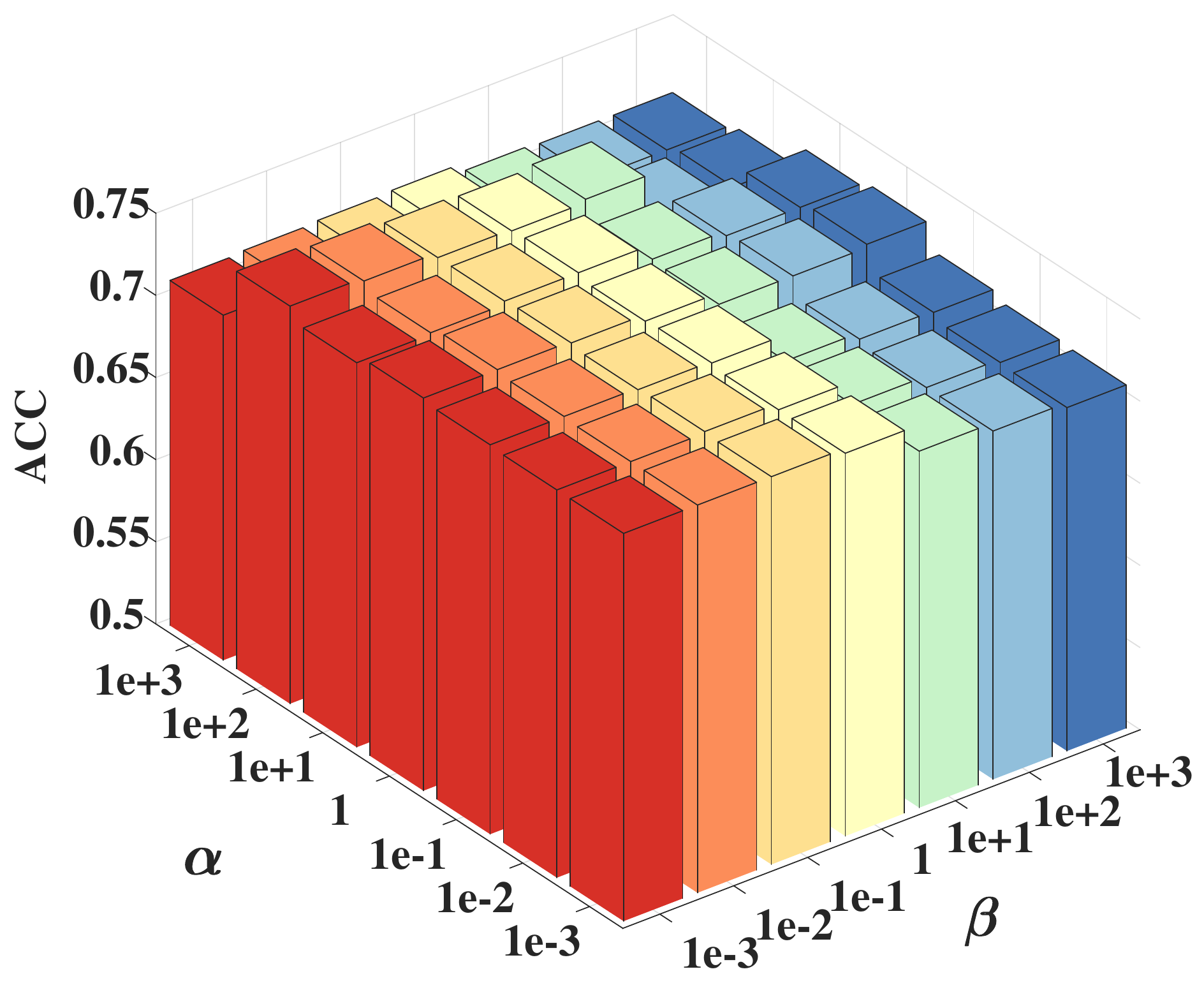}
\caption{Citeseer}
\label{linmingdu_Citeseer}
\end{subfigure}
\begin{subfigure}{0.23\textwidth}
\includegraphics[width=1\textwidth]{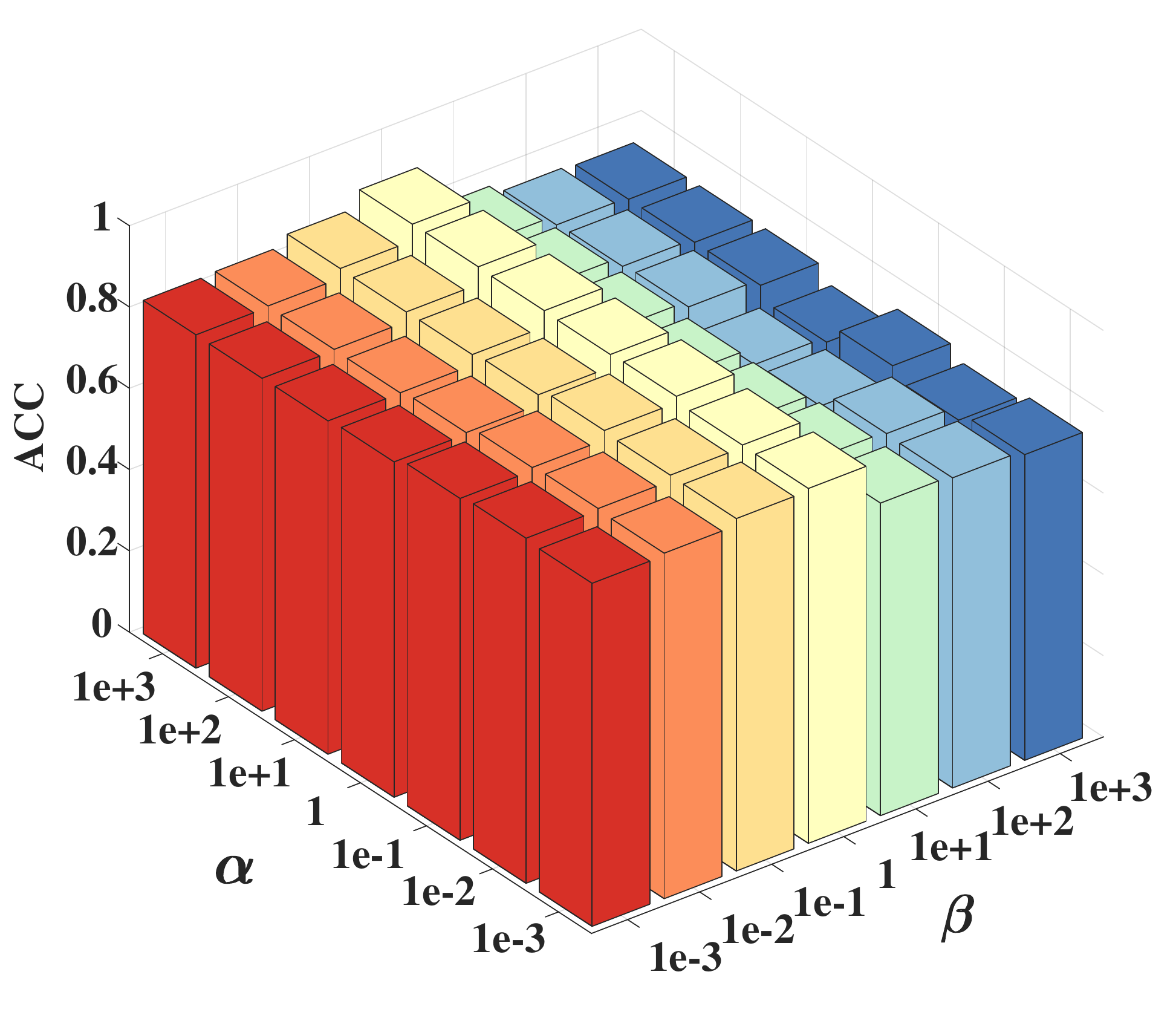}
\caption{BlogCatalog}
\label{linmingdu_BlogCatalog}
\end{subfigure}
\caption{Sensitivity analysis of parameters $\alpha $ and $\beta $.}
\label{linmingdu}
\end{figure}

Moreover, we analyze the sensitivity of crucial parameters on datasets Citeseer and BlogCatalog. Specifically, we test the influence of the constraint weight parameters $\alpha $ and $\beta $ in Eq.\eqref{loss_function} and change them from $1e - 3$ to $1e + 3$, where $\alpha $ and $\beta $ are related to feature graph and semantic graph respectively. For better visualization, we plot a histogram of the ACC values with respect to different parameters, as shown in Fig. \ref{linmingdu}.

We can observe that with the increase of $\alpha $ and $\beta $, the ACC values generally rise first and then drop. When the parameters are in the preset range, ML-SGNN is basically stable, which shows that the method is robust to the parameters $\alpha$ and $\beta$.
We also find that in Fig. \ref{linmingdu_BlogCatalog}, the performance has a sharp drop when $\beta>1$ on BlogCatalog, while it is relatively stable on Citeseer.
The possible reason is that in this case, only the original topology graph and the feature graph are used while the semantic graph is ignored, so ML-SGNN degenerates into a vanilla GNN model. This significant performance degradation also shows the effectiveness of module aggregation learning. 
Moreover, performance is improved relative to parameters $\alpha $ and $\beta $ approaching 0 when the parameters are within a reasonable range, which explains the effectiveness of the sparse-inducing ${{l}_{2,1}}$ regularizer constraint in Section \ref{sec3.4.2}.
In addition, the sensibility of $\alpha $ and $\beta $ behaves varyingly on different datasets, which shows that the importance of feature graph and semantic graph is different and needs further evaluation.  


\section{Conclusion}\label{sec5}
In this work, we have proposed a framework named ML-SGNN for semi-supervised classification tasks. It explores the complex interactions in semantics by generating a semantic graph to preserve the global graph structure, and considers the strategy of similarity evaluation by adaptively merging various measures for better embeddings. Extensive experiments have shown that our proposed robust semantic graph learning and multi-measure learning have a significant impact on preserving the global structure of the model and node embedding optimization. For future work, we plan to utilize semantic graph learning and multi-measure learning for improving heterogeneous node embedding.


\section{Acknowledgements}
This work is supported by Excellent Dissertation Cultivation Funds of Wuhan University of Technology under Grant 2021III030JC.

\bibliographystyle{model1-num-names}
\bibliography{bibfile2}

%


\printcredits

\appendix

\section{SUPPLEMENT}
In the supplement, we provide the websites of baselines and datasets of all experiments involved in this paper. For reproducibility, we also provide the specific experimental environment and the hyperparameter values of all experiments.


\begin{table}
\centering
\setlength{\tabcolsep}{1.1mm}
\caption{Parameter settings}
\begin{tabular}{c|c|cccccc}
\hline
dataset            & L/C &  lr  & $\varpi $ & nhid1 & nhid2 & $\alpha$ & $\beta$ \\ \hline
\multirow{3}{*}{Citeseer}   & 20  & 5e-4 &   5e-3    &  768  &  128  &   100    &  0.001  \\
& 40  & 5e-4 &   5e-3    &  768  &  128  &    10    &  0.001  \\
& 60  & 5e-4 &   5e-3    &  768  &  128  &    10    &  0.01   \\ \hline
\multirow{3}{*}{UAI2010}   & 20  & 5e-4 &   5e-4    &  512  &  128  &    1     &  0.01   \\
& 40  & 5e-4 &   5e-4    &  512  &  128  &   0.1    &  0.01   \\
& 60  & 5e-4 &   5e-4    &  512  &  128  &   0.1    &  0.01   \\ \hline
\multirow{3}{*}{ACM}     & 20  & 1e-4 &   6e-4    &  768  &  256  &  0.001   &  0.001  \\
& 40  & 1e-4 &   5e-4    &  768  &  256  &    1     &  0.001  \\
& 60  & 5e-4 &   5e-4    &  768  &  256  &    1     &  0.001  \\ \hline
\multirow{3}{*}{BlogCatalog} & 20  & 3e-4 &   1e-5    &  768  &  128  &   1000   &  0.001  \\
& 40  & 5e-4 &   1e-5    &  768  &  128  &   100    &  0.001  \\
& 60  & 3e-4 &   1e-5    &  768  &  128  &   100    &  0.001  \\ \hline
\multirow{3}{*}{Flickr}    & 20  & 5e-4 &   1e-5    &  512  &  128  &   0.1    &    1    \\
& 40  & 5e-4 &   1e-5    &  512  &  128  &   0.1    &   10    \\
& 60  & 5e-4 &   1e-5    &  512  &  128  &   0.1    &   10    \\ \hline
\multirow{3}{*}{CoraFull}   & 20  & 1e-3 &   5e-4    &  512  &  32   &  0.001   &  0.001  \\
& 40  & 1e-3 &   5e-4    &  512  &  32   &  0.001   &  0.001  \\
& 60  & 1e-3 &   5e-4    &  512  &  32   &  0.001   &  0.001  \\ \hline
\end{tabular}
\label{parameter}
\end{table}

\subsection{Experiments Settings}
All experiments are conducted with the following setting:
\begin{itemize}
\item CPU: AMD Ryzen 7 3750H @2.30 GHz
\item GPU: GeForce GTX 2060
\item Software versions: Python 3.8; Pytorch 1.7.1; Numpy 1.19.2; SciPy 1.6.0; NetworkX 2.4; Scikit-learn 0.24.1
\end{itemize}

\subsection{Implementation Details}
The codes of ML-SGNN are based on PyTorch version of Graph Convolution Networks. For reproducibility, the code are publicly available\footnote[1]{\url{https://github.com/landrarwolf/ML-SGNN}} and model hyperparameters are listed in Table \ref{parameter}.

\subsection{Sensitivity Analysis on a Wider Scale}
\begin{figure}
\centering
\includegraphics[width=0.7\linewidth]{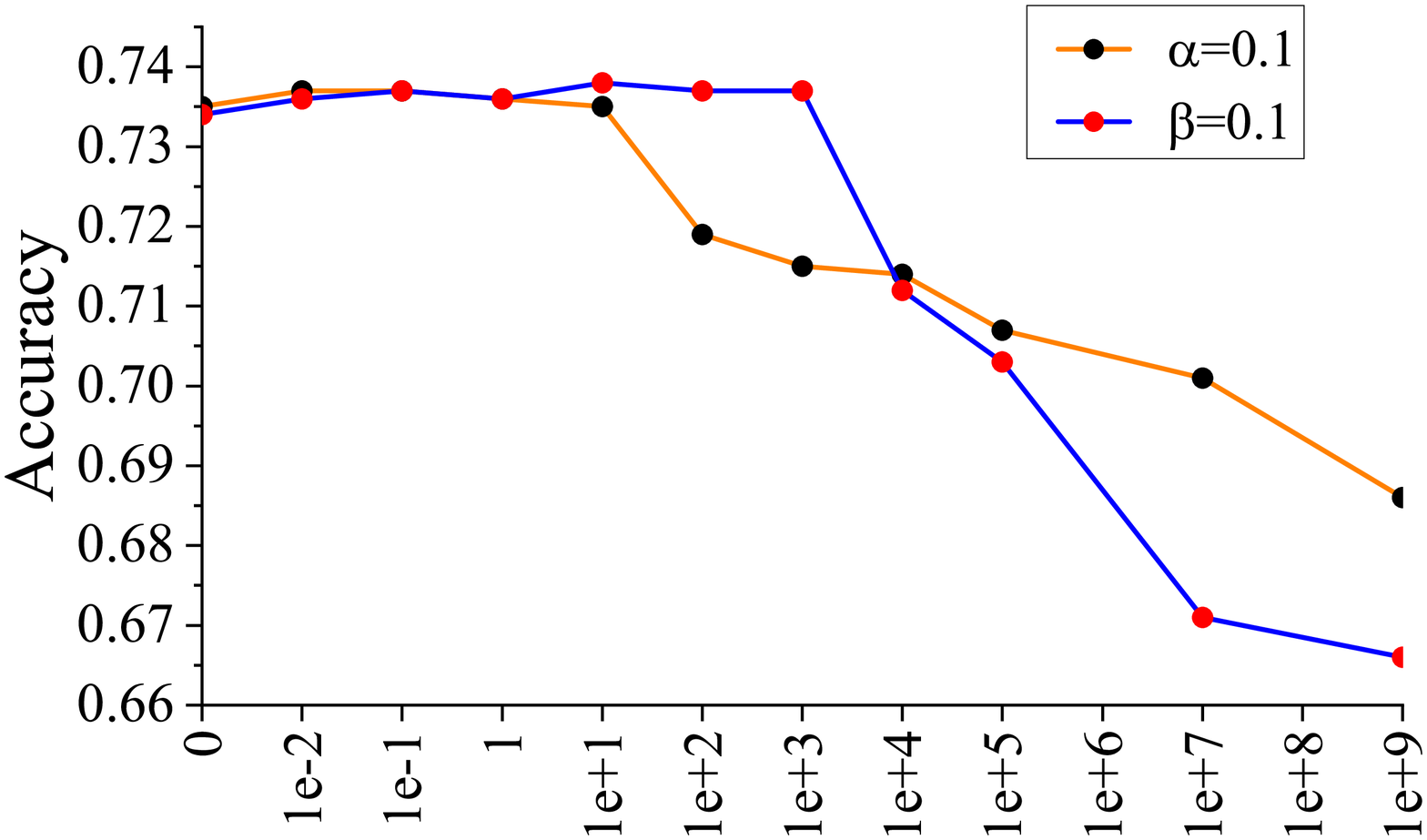}
\caption{Analysis of hyperparameter $\alpha$ and $\beta$ on a larger scale.}
\label{fig:largerscale}
\end{figure}

In Section \ref{para}, we result that the parameters $\alpha$ and $\beta$ have less effect on ACC within a certain range, which shows that our ML-SGNN is robust to the two hyperparameters. To further verify whether this conclusion is still valid in a wider range, we extend the experiment on the effect of hyperparameters on ACC, expanding the value by several orders of magnitude. Specifically, we test the classification accuracy of ML-SGNN with a label rate of 20L/C on the Citeseer dataset with the changes of hyperparameters $\alpha$ and $\beta$. The results are shown in Fig. \ref{fig:largerscale}.

The two lines represent that when one parameter is fixed at 0.1, the model performance differs with the change of the other parameter. It is worth noting that we widen the parameter range to [0, 1e + 9]. It can be found that:
\begin{itemize}
\item If $\alpha$ or $\beta$ is small enough (we assume the value is 0), then $ \mathcal{L}={{\mathcal{L}}_{0}} $in Equation 16, i.e., the regularization term is ignored. The trained model at this time have a certain degree of overfitting, resulting in a small decline in performance.
\item If $\alpha$ or $\beta$ is large enough (we assume values is 1e + 4 and larger), like other network models, the weights of the regularization term will be overly amplified, resulting in severe limitation of network training.
\end{itemize}
These similar patterns exist on a wider range of datasets. Thus, our ML-SGNN has robustness to hyperparameters $\alpha$ and $\beta$ in a less stringent range of values. Even so, like other neural network models, hyperparameters that are not within a reasonable range can still affect performance.
\end{sloppypar}
\end{document}